\documentclass{article}

\usepackage[preprint]{neurips_2019}

\usepackage[utf8]{inputenc} 
\usepackage[T1]{fontenc}    
\usepackage{hyperref}       
\usepackage{url}            
\usepackage{booktabs}       
\usepackage{amsfonts}       
\usepackage{nicefrac}       
\usepackage{microtype}      
\usepackage{color}
\usepackage{amsmath}
\usepackage{algorithmic}
 
\usepackage{hyperref}

\usepackage{wrapfig}
\usepackage{subcaption}
\usepackage{subcaption} 
\usepackage[ruled]{algorithm2e}
\usepackage{amssymb,amsmath}
\usepackage{amsthm}
\usepackage{bm}
\usepackage{dsfont}

\def\EXP{\mathbb{E}}

\newcommand{\ent}{\mathbb{H}}

\usepackage{graphicx}
\graphicspath{{Plots/}}

\title{Marginalized State Distribution \\
Entropy Regularization in Policy Optimization}

\author{%
Riashat Islam\\
McGill University, Mila\\
School of Computer Science\\
riashat.islam@mail.mcgill.ca\\
\And
Zafarali Ahmed\\
McGill Univeristy, Mila\\
School of Computer Science\\
zafarali.ahmed@mail.mcgill.ca\\
\AND
Doina Precup\\
McGill University, Mila\\
School of Computer Science\\
dprecup@cs.mcgill.ca\\
}

\begin{document}

\maketitle

\begin{abstract}
Entropy regularization is used to get improved optimization performance in reinforcement learning tasks. A common form of regularization is to maximize policy entropy to avoid premature convergence and lead to more stochastic policies for exploration through action space. However, this does not ensure exploration in the state space. In this work, we instead consider the distribution of discounted weighting of states, and propose to maximize the entropy of a lower bound approximation to the weighting of a state, based on latent space state representation. We propose entropy regularization based on the marginal state distribution, to encourage the policy to have a more uniform distribution over the state space for exploration. Our approach based on marginal state distribution achieves superior state space coverage on complex gridworld domains, that translate into empirical gains in sparse reward 3D maze navigation and continuous control domains compared to entropy regularization with stochastic policies.

\end{abstract}

\section{Introduction}
A key ingredient of a successful reinforcement learning algorithm is sufficient exploration of the environment. This is particularly important when rewards that the environment provides are sparse. In policy optimization, entropy regularization is added to the objective to prevent the policy from prematurely converging to a deterministic policy \citep{mhiha2c,sac} leading to improved optimization performance \citep{understanding_entropy}. Simply regularizing by the policy entropy induces a stochastic policy via the action space. However, exploration in action space does not imply exploration in state space: This policy would perform a random walk in environments with sparse rewards.

In contrast, an effective exploration policy should also seek to maximize coverage of the state space. Most recently, \citet{kakade_entropy} proposed a framework that maximizes exploration by maximizing the entropy of  the discounted stationary state distribution induced by a policy, $d_\pi$. While the guarantee is useful, their technique relies on having access to an (approximate) model and requires state discretization in continuous state space environments. An alternative is to compute the normalized discounted weighting of a state, or commonly known as the discounted future state distribution. However, as we discuss in later sections, this is also difficult to compute as it requires estimating the probability of a state with a $(1-\gamma)$ probability. Additionally, estimating the entropy of the discounted occpancy measure would require separate density estimates, while the entropy of stationary distribution requires knowledge of the environment transition dynamics. 

In this work, we propose a practical model-free approach to estimating the distribution over discounted weighting of states, for entropy regularization with the state distribution. Since the normalized discounted weighting is difficult to compute, we instead estimate the marginal state distribution that measures the probability of being in state s at time t dependent on the policy parameters, $d_{t,\pi}(s) = P_\pi(s_t = s)$. We propose to use a lower bound approximation to $d_{t,\pi}(s)$ based on latent space representation of a state to compute $d_{z, t,\pi}(s)$ and compute the entropy based on the latent representation $\ent( d_{z, t,\pi}(s) ) $ for entropy regularization. We hypothesize that maximizing the entropy of marginal state distribution can be an effective exploration method that maximizes coverage of the state space, and propose a computationally feasible algorithm based on a variational approach.

Effective exploration in sparse reward domains is a challenging problem and several heuristics have been proposed to incentivize the agent to visit unseen regions of the state space. These methods introduce explicit reward shaping bonuses for exploration \citep{Bellmare_Count,pathak2017curiosity,bellemare_density,machado2018count} but do not explicitly \emph{regularize} policies to maximize an exploration objective. Specifically, there is no gradient of the exploration bonus with respect to the policy parameters. In policy optimization, regularization is a simple way to extract specific behaviours from policies \citep{bachman2018vfunc,infobot,eysenbach2019diversity,pong2019skew}. 

In this work, we investigate the following question: \textit{Can maximizing entropy of marginal state distribution for regularization be useful as a simple exploration strategy?} Our contributions are:
\begin{itemize}
    \item To justify the maximization of state space coverage, we first study the use of discounted state distribution in episodic settings, as an effective entropy regularizer, $\ent (d_{\pi})$. These results complements \citet{kakade_entropy}. %
    \item We discuss why computing the normalized discounted weighting of a state, or the discounted future state distribution is difficult in practice. We then propose an approximation to the discounted weighting of a state, by instead computing a lower bound representation of the discounted weighting. We show that we can tractably compute the discounted per state weighting, or equivalently the marginal state distribution, based on latent state representation. 
    \item We then propose to use the entropy of the marginal state distribution, $\ent ( d_{t, \pi}(s))$ for entropy regularization with the approximation $\ent ( d_{z, t, \pi}(s))$ in policy optimization. By introducing a tractable algorithm, we show that it induces a policy that maximize state space coverage and study other qualitative and quantitative behaviours in Grid Worlds and maze tasks.
    \item In more complicated domains, we demonstrate that maximum marginal state entropy regularization objective can induce an effective exploration strategy. In particular, it shows improved performance across a wide range of tasks, including challenging maze navigation domains with sparse reward, partially observable environments, and even dense reward continuous control tasks.
\end{itemize}

\section{Preliminaries}
\label{sec:background}
In reinforcement learning, we aim to find a policy, $\pi(a|s)$, that maps a state $s$, to a distribution over actions, $a$, such that it maximizes the cumulative discounted return: $J(\pi) = \EXP_{s\sim d_{\pi}(s), a \sim \pi(a|s)}\big[\sum_{t=0}^{\infty} \gamma^{t} r_t\big]$ where $d_{\pi}(s)$ is the stationary state distribution of the policy, $\gamma$ is a discount factor favouring immediate rewards and $r_t$ is a reward at time $t$ in the trajectory. In the infinite horizon setting, the state distribution is the normalized distribution given by $d_{\pi}(s) = (1 - \gamma) \sum_{t=1}^{\infty} \gamma^{t} d_{t, \pi}(s)$. Here $d_{t, \pi}(s) = p_\pi(s_t)$ is the per-step state distribution induced by a policy $\pi$. In policy gradient methods the goal is to find a parameterized policy, $\pi_\theta$, that maximizes the discounted cumulative return, $J(\pi)$. A practical algorithm is given by following the policy gradient, $\nabla_\theta J(\pi)$ \citep{sutton}. 
Maximum entropy based objectives \citep{schulman2017equivalence,understanding_entropy} augment the reward with policy entropy penalty, $r_t + \ent (\pi_{\theta})$, to help with exploration by avoiding premature convergence of $\pi_\theta$ to a deterministic policy. The entropy regularized objective is: $\tilde{J}(\theta) = \EXP_{\pi_{\theta}} \big[\sum_{t=0}^{\infty} \gamma^{t} r_t + \lambda_\pi \ent (\pi_{\theta})  \big]$ where $\lambda_\pi$ determines the relative importance of the entropy term against the reward. The gradient of $\tilde{J}$ is given by $\nabla_{\theta}\tilde{J}(\theta) = \EXP_{s\sim d_{\pi}(s), a\sim \pi} [\nabla_{\theta} \log \pi_{\theta}(a|s) Q^{\pi_\theta,\lambda_\pi}(s,a) + \lambda_\pi \nabla_{\theta} \ent (\pi_\theta)]$,
where $Q^{\pi,\lambda_\pi}(s,a)$ is the action-value function describing the expected discounted entropy augmented return by executing action $a$ in state $s$ and then acting on-policy by sampling $a\sim \pi(\cdot|s)$. 

One could imagine that in addition to maximizing $\ent(\pi)$, we can also maximize the entropy of the state distribution, $\ent (d_\pi)$. 
The main focus of our work is to propose a tractable approach to maximizing this entropy via the entropy of the per-step state distribution, $ \ent( d_{t, \pi_{\theta}})$ (Section~\ref{sec:approach:basic}) and a variational approximation (Section~\ref{sec:approach:tractable_approximation}).

\section{Approach: Entropy Regularization with Marginal State Distribution}
\label{sec:approach}

\subsection{Marginal State Distribution}
\label{sec:marginal_state_dist_definition}

To understand the meaning of the marginal state distribution, let us first recall that the exact solution for the policy evaluation equation in vector form is given by $\mathbf v_{\pi}(s) = (\mathbf I - \gamma \mathbf P_{\pi})^{-1} \mathbf r_{\pi}$, where $\mathbf r_{\pi}$ is the reward and $\mathbf P_{\pi}$ defines the dynamics of the Markov chain. The inverse of $\mathbf I - \gamma \mathbf P_{\pi}$ can be written as $(\mathbf{I} - \gamma \mathbf P_{\pi})^{-1} = \sum_{t=0}^{\infty} (\gamma \mathbf P_{\pi})^{t}$, also known as the Neumann series, and each row of $P_{\pi}^{t}$ is the distribution over next states $t$ steps in the future. Therefore, we can write the probability of the agent being in state $s'$, after $t$ steps in the future as $\mathbf P_{\pi}^{t}(s, s') = P(s_t = s' \mid s_0 = s)$, and each entry the inverse matrix be written as  $(\mathbf I - \gamma \mathbf P_{\pi})^{-1}(s, s') = \sum_{t=0}^{\infty} \gamma^{t} P_{\pi} (s_t = s' \mid s_0 = s)$.

Furthermore, as in \citep{sutton}, the policy gradient discounted objective given an initial starting state distribution $\alpha$ is given by $J(\pi) = \alpha^{T} \mathbf v_{\pi}   = \mathbf \alpha^{T} (\mathbf I - \gamma \mathbf P_{\pi})^{-1} r_{\pi} $. Given the distribution of initial states, we can define the \textit{discounted weighting of states} as $d_{\pi} = \mathbf \alpha^{T} (\mathbf I - \gamma \mathbf P_{\pi})^{-1}$, such that the equivalent definition of policy gradient objective is $J(\pi) = \mathbf d_{\pi}^{T} \mathbf r_{\pi}$. The discounted weighting of states can further be written as a recursive expression, similar to policy evaluation equations, where $d_{\pi}$ plays the role of value function $\mathbf d_{\pi}^{T} = \alpha^{T} \sum_{t=0}^{\infty} (\gamma P_{\pi})^{t} = \alpha^{T} + \gamma d_{\pi}^{T} P_{\pi}$. The recursive expression for the discounted weighting of states can therefore be written as an expectation
\begin{equation}
\label{eq:discounted_weighting_states}
    d_{\pi}(s) = \sum_{s_0} \alpha (s_0) \sum_{t=0}^{\infty} \gamma^{t} P_{\pi} (s_t = s \mid s_0 = s_0) = \EXP{_{\alpha}} \Big[  \sum_{t=0}^{\infty} \gamma^{t} P_{\pi} (s_t = s)  \Big]
\end{equation}
Note that the value function is also defined given a starting state or as an expectation where the states are drawn from a starting state distribution $\alpha$, since it is $v_{\pi}(s) = \EXP \Big[  \sum_{t=0}^{\infty} \gamma^{t} r(s_t) \mid s_0 = s \Big] = \EXP{_{\alpha} \Big[  \sum_{t=0}^{\infty} \gamma^{t} r(s_t)  \Big]}$. We can therefore write the policy gradient objective with an equivalent expectation w.r.t to the starting state distribution $\alpha$, given by $J(\pi) = \EXP{_{\alpha}} \Big[  v_{\pi}(s) \Big]$. Note that since both the discounted weighting of states, given in equation \ref{eq:discounted_weighting_states} and the value function are defined with $\EXP{_{\alpha}}$, we can further write the joint expectation, where we denote $\tilde{v}_{\pi}(s)$ as the value function with augmented rewards, given by $P_{\pi} (s_t = s)$. We define $P_{\pi}(s_t=s)$, the probability of being in state s at time t as the marginal discounted weighting of states. Further from equation \ref{eq:discounted_weighting_states}, let us denote $P_{\pi}(s_t=s) = d_{t, \pi}(s)$, as the marginal state distribution, we can therefore write the discounted weighting of states as $d_{\pi}(s) = \sum_{s_0} \alpha (s_0) \sum_{t=0}^{\infty} \gamma^{t} d_{t, \pi} (s)$.

\begin{equation}
\begin{split}
\label{eq:reward_augmented_value_functions}
\tilde{v}_{\pi}(s) &= \EXP{_{\alpha}} \Big[  \sum_{t=0}^{\infty} \gamma^{t} r(s_t) + \sum_{t=0}^{\infty} \gamma^{t} P_{\pi}(s_t = s)  \Big]\\
&= \EXP{_{\alpha}} \Big [  \sum_{t=0}^{\infty} \gamma^{t} [ r(s_t) + P_{\pi}(s_t=s) ] =  \EXP{_{\alpha}} \Big [  \sum_{t=0}^{\infty} \gamma^{t} [ r(s_t) + d_{t, \pi}(s) ] \Big]
\end{split}
\end{equation}

 The policy gradient objective, given by value functions $\tilde{v}_{\pi}(s)$ with augmented rewards is $\tilde{J} (\pi) = \EXP{_{\alpha}} \Big[ \tilde{v}_{\pi}(s) \Big]$, where the augement rewards are given by $d_{t, \pi}(s) = P_{\pi} (s_t = s)$. The intuition for using such augmented rewards is similar to count-based occupancy measures for example, where the agent is provided an exploration bonus for states that are visited less often. However, the key difference is that here the occupancy measure is dependent on the policy $P_{\pi}(s_t = s)$ directly, as it derives from the discounted weighting of future states dependent on the policy $\pi$. 

However, from equation \ref{eq:discounted_weighting_states} note that $d_{\pi}(s)$ is not in fact a distribution, since the rows of $(\mathbf I - \gamma \mathbf P_{\pi})^{-1}$ do not sum to 1, but instead would sum to $\frac{1}{(1 - \gamma)}$. Therefore, to consider the distribution of discounted weighting of states, we would need to consider the normalized variant that we denote by $d_{\gamma, \pi}(s) = (1 - \gamma) d_{\pi}(s) = (1-\gamma) \sum_{t=0}^{\infty} \gamma^{t} d_{t, \pi}(s)$. We would therefore also need to consider the normalized variant of the marginal discounted weighting of states, which we define as the \textit{marginal state distribution} given by $d_{\gamma, t, \pi}(s)$
\[  d_{\gamma, t, \pi} = ( 1 - \gamma) d_{t, \pi}(s)  \]. This means that instead of augmenting rewards in $\tilde{v}_{\pi}(s)$ with $P_{\pi}(s_t=s)$, we would instead need to augment it with $(1-\gamma) P_{\pi}(s_t=s)$ or equivalently denoting it as $d_{\gamma, t, \pi}$. This however is quite difficult in practice since in order to estimate $d_{\gamma, t, \pi}$ we would need to estimate the probability of being in a state with $(1-\gamma)$. In other words, for the normalized discounted weighting of states $d_{\gamma, \pi}(s)$ defined above, we would now need to sample upon entering the next state, from the geometric distribution of $(1 - \gamma) \gamma^{t}$ to decide whether to terminate from the trajectory prematurely with a probability of $(1-\gamma)$ or not.

\subsection{Marginalized State Distribution Entropy Regularization}
\label{sec:approach:basic}
Instead of augmenting rewards with $\mathbf P_{\pi}(s_t=s)$, we can instead consider different entropies.
Instead of value functions augmented with rewards $P_{\pi}(s_t=s)$, we can instead consider entropy augmented rewards, similar to entropy regularized policies $\ent (\pi)$ \citep{mhiha2c}. We can therefore consider regularized value functions with the entropy of the marginal state distribution $\ent (P_{\pi}(s_t = s))$ or equivalently $\ent (d _{t, \pi}(s))$. We can re-write equation \ref{eq:reward_augmented_value_functions} with entropy augmented value functions, with weighting $\lambda$ and taking the expectation, $\EXP{_{\pi}}$ over all states along the trajectory, to get the policy gradient objective with parameterized policies $\pi_{\theta}(a,s)$ given by 
\begin{equation}
\label{eq:objective_regularized}
    J(\theta) = \EXP{_{\pi_{\theta}}} \Big[ \sum_{t=0}^{\infty} \gamma^{t} [  r(s_t)  + \lambda \ent( d_{t, \pi_{\theta}}(s) ) \Big] = \EXP{_{\pi_{\theta}}} \Big[ \sum_{t=0}^{\infty} \gamma^{t} [  r(s_t)  - \lambda \log P_{\theta}(s_t = s) \Big] 
\end{equation}

We can therefore regularize policy gradient objective with the entropy of the marginal state distribution, which we denote by $\ent ( P_{\theta}(s) )$. As with policy entropy regularization, we  augment the reward, $r_t$, with the entropy of the marginal state distribution, $\ent ( P_{\theta}(s_t) )$, given by $r_t = r_t + \lambda_\pi \ent(\pi_{\theta}) + \lambda_s \ent ( P_{\theta}(s_t) )$ where $\lambda_s,\lambda_\pi\geq 0$ are weighting terms.\footnote{Observe that unlike other reward bonuses, both $\ent(\pi_\theta)$ and $\ent(P_{\pi_\theta})$ depend on the policy parameters, $\theta$, and will therefore have non-zero gradients with respect to $\theta$.} For simplicity, we will focus on the second term in the following derivations. Similar to count based exploration \citep{Bellmare_Count, bellemare_density}, this reward term would incentivize policies which increase $\ent (P_{\theta}(s))$ and therefore the diversity of states visited. 


Using this reward bonus results in a slightly different policy gradient objective, ${\tilde{J}(\theta) = \EXP_{\pi_{\theta}} \Bigl[\sum_{t=0}^{\infty} \gamma^{t} r(s_t, a_t) + \lambda_{s} \ent ( P_{\theta}(s_t)) )  \Bigr]}$ in which the gradient of the entropy of the marginal state distribution, $\ent ( P_{\theta}(s_t))$ can act as regularizer, leading to the state entropy regularized policy gradient update: 
\begin{equation}
\label{eq:state_entgrad}
\nabla_{\theta} \tilde{J}(\theta) = \EXP_{\pi_{\theta}}  \Bigl[ \nabla_{\theta} \log \pi_{\theta}(a_t | s_t) Q^{\pi_{\theta},\lambda_s}(a_t, s_t) + \lambda_{s} \nabla_{\theta} \ent ( P_{\theta}(s_t) )     \Bigr]
\end{equation}
where $Q^{\pi_{\theta},\lambda_s}$ is the cumulative discounted augmented reward from state $s_t$ taking action $a_t$ and then acting on-policy according to $a\sim\pi_\theta$. Derivation of the marginal state entropy regularized policy gradient is given in Appendix~\ref{sec:appendix:pg_derivation}.  

However, note that in equations \ref{eq:objective_regularized} and \ref{eq:state_entgrad}, we have defined the objective w.r.t to the unnormalized version of the marginal discounted weighting of states. In other words, to consider the marginal state distribution, we would need $\ent (d_{\gamma, t, \pi}(s))$ instead of $\ent ( d_{t, \pi}  )$. In other words, we would need the normalized marginal probability of state $(1 - \gamma) P_{\theta}(s_t = s)$ to turn it into a distribution. This is intractable in practice for two reasons : (a) we cannot estimate probability with a $(1 - \gamma)$ normalization factor since it means to compute the marginal probability with a probability of $(1 - \gamma)$; (b) for continuous state spaces and in case of function approximation, we cannot get exact estimates of the marginan probability of being in state s at time t, $P_{\theta}(s_t = s)$ dependent on the policy $\pi_{\theta}(a,s)$. Computing the marginal state distribution $P_{\theta}(s_t)$ or $d_{t, \pi}(s)$ for time-step $t$ therefore requires an intractable integral across all the time-steps of a trajectory. 

In the next section, we discuss how we can compute an approximation to the marginal state distribution $d_{\gamma, t, \pi}(s)$, which is dependent with changes in policy $\pi_{\theta}$, by introducing a variational approximation \citep{vae_kingma} $q_{\theta}(z_t|s_t)$.

\subsection{Approximation to Marginal State Distribution Entropy}
\label{sec:approach:tractable_approximation}

In practice, it is difficult to compute the normalized marginal state of a probability, $p_{\theta}(s)$ dependent on policy parameters $\theta$.
We introduce use a variational approximation $q_\theta(z | s_t)$ to $p_{\theta}(s_t)$. In particular, we assign a probability distribution over each state $p_{\theta}(s_t)$ by using an encoder to map each state to a latent representation, $z$, of that state $q_\theta(z|s_t)$. This can be achieved by defining a policy network which outputs $p(a,z|s)$. We can decompose the output of the policy as $p_{\theta}(a,z|s) = \pi_{\theta}(a|s) q_{\theta}(z|s)$. We define a policy function, which in addition to mapping states to actions, also maps states to a latent representation $z$. Informally, this means that for each state $s_t$ visited at time t under the current policy parameterization $\pi_{\theta}(a,s)$, we compute the corresponding marginal probability of being in that state. However, since $P_{\theta}(s_t)$ cannot be computed exactly for the marginal state distribution, we instead compute a lower bound approximation, ie, a latent state representation and compute the marginal probability distribution in the latent space $q_{\theta}(z\mid s_t)$. 
Using the variational approximation $q_{\theta}(z \mid s)$, we can now re-define the discounted weighting of states, based on the latent representation that we denote by $d_{z, \pi}(s)$, which is a lower bound approximation to $d_{\pi}(s)$ as previously given in equation \ref{eq:discounted_weighting_states}

\begin{equation}
d_{\pi}(s) = \sum_{s_0} \alpha (s_0) \sum_{t=0}^{\infty} \gamma^{t} P_{\pi} (s_t = s \mid s_0 = s_0) \geq \sum_{s_0} \alpha (s_0) \sum_{t=0}^{\infty} \gamma^{t} q_{\pi} (z_t \mid s_t) = d_{z, \pi}(s)
\end{equation}

Therefore, based on the latent representation, we can again define the normalized latent state distribution relating to the marginal latent state distribution as $d_{z, \pi}(s) = (1 - \gamma) \sum_{t=0}^{\infty} \gamma^{t} d_{t, z, \pi}(s)$. The marginal state distribution is therefore given by $d_{t, z, \pi}(s) = q_{\pi} (z \mid s)$.

In marginal state entropy regularized policy gradients with parameterized policies $\pi_{\theta}(a,s)$, we therefore require a tractable approximation to computing the entropy of marginal state distribution $\ent(d_{t, \theta})$ or equivalently defined as $\ent (P_{\theta}(s_t))$. The variational entropy, $\ent (q_{\theta}(z|s))$, gives an approximation to the marginal state distribution entropy, $\ent (P_{\theta}(s))$, and therefore we can maximize this approximation instead. This approximation, however, may have drawbacks from a theoretical standpoint, since $q_{\theta}(z|s)$ and $P_{\theta}(s)$ may decouple the assigned distributions over a given state $s_t$. However, we find that maximizing the lower bound $q_{\theta}(z|s)$ still maximizes $P_{\theta}(s)$ and leads to benefits with exploration in practice. We argue that this approach still provides a suitable approximation to the policy dependent marginal state distribution $P_{\theta}(s)$, which may otherwise be difficult to compute exactly. Using the variational approximation, we get a lower bound to the objective in equation\ref{eq:objective_regularized}.

\begin{equation}
  \small J(\theta) = \EXP{_{\pi_{\theta}}} \Bigl[  \sum_{t=0}^{\infty} \gamma^{t} [  r(s_t)  + \lambda \ent( P_{\theta}(s_t) )    \Bigr]
    \small \geq \EXP{_{\pi_{\theta}}} \Bigl[  \sum_{t=0}^{\infty} \gamma^{t} [  r(s_t)  + \lambda \ent( q_{\theta} (z_t \mid s_t) )    \Bigr]   
\end{equation}


From the above equation, we get the following approximation to the policy gradient with the variational state distribution entropy $\ent(q_{\theta}(z|s))$, where the term $\ent ( q_\theta(z | s_t) ) $ is the variational marginal entropy of $\ent (p_{\theta}(s))$ and acts as a regularizer in the policy gradient update since the entropy term per state relates to the changes in the policy.
\begin{equation}
\label{eq:state_entgrad_approx}
\nabla_{\theta} \tilde{J}(\theta) = \EXP_{\pi_{\theta}}  \Bigl[ \nabla_{\theta} \log \pi_{\theta}(a_t | s_t) Q^{\pi_{\theta}}(a_t, s_t) + \lambda_{s} \nabla_{\theta} \ent ( q_\theta(z_t | s_t)) )     \Bigr]
\end{equation}


\paragraph{Algorithm:} Our algorithm is summarized in Algorithm 1. As described in Section~\ref{sec:approach:tractable_approximation}, we use a policy architecture that outputs both the action probabilities and the encoded latent state representation of states $s$, denoted by $p(z|s)$. 

\begin{algorithm}[!htb]
\caption{Regularization with Entropy of Marginal State Distribution $\ent (P_{\theta}(s))$ }
\begin{algorithmic}
\label{algo:training}
\REQUIRE ~~ A policy $\pi_{\theta}\!\left(a, z \mid s\right) = q_{\theta}\!\left( z \mid s\right) p_{\theta}(a\mid s)$ and $\lambda_s$, $\lambda_\pi$ regularization coefficients.
\REQUIRE ~~ The number of episodes, $E$ and update interval, $N$.
\FOR{$e=1$ to $E$}
\STATE{Take action $a_{t}$,get reward $r_{t}$ and observe next state $s_{t+1}$}
\STATE{Store tuple ($s_t,a_t, r_{t+1}, s_{t+1}$) as trajectory rollouts or in replay buffer $\mathcal{B}$}\\
        \uIf{$\mod(e, N)$}{
        Update policy parameters $\theta$ following \textit{any} policy gradient method, according to\\
        $\nabla_{\theta} \tilde{J}(\theta) = \Bigl[ \nabla_{\theta} \log \pi_{\theta}(a_t | s_t) Q^{\pi_{\theta}}(a_t, s_t) + \lambda_{\pi} \ent ( \pi_{\theta} ( \cdot | s_t ) ) + 
        \lambda_{s} \nabla_{\theta} \ent ( q_{\theta}(z_t | s_t ))     \Bigr]$
        }
\ENDFOR 
\end{algorithmic}
\end{algorithm}

\section{Experiments}
\label{sec:experiments}
In this section, we demonstrate the usefulness of state entropy regularization, specifically marginal state entropy regularization, for exploration. We provide a detailed experimental setup for each section as well as reproducibility efforts in Appendix~\ref{sec:ref:detailed_experimental_setup}. In experimental results below, we denote \emph{MaxEntPolicy} for regularization with $\ent (\pi_{\theta})$ only and \emph{MaxEntState} for regularization with $\ent (p_{\theta}(s))$ in addition with policy entropy.\footnote{Unless specified, all our baselines use $\lambda_\pi=0.1$.} 

The goal of our experiments is to demonstrate that a simple idea and regularization method for marginal state distribution entropy maximization can lead to better performance on a range of sparse reward tasks. Firstly, in Section~\ref{sec:res:hazan}, we verify the hypothesis from \citet{kakade_entropy} that maximizing the entropy of the stationary state distribution $\ent (d_{\pi})$ can lead to faster learning when computing the \textit{exact policy gradient}. In Section~\ref{sec:res:state_space_coverage} we show that our tractable approximation to maximizing the marginal state entropy, $\ent (p_{\theta}(s))$, can induce a policy which will maximize state coverage, leading to better exploration of the state space in a variety of gridworlds, compared to maximizing policy entropy. Finally we show that the increased state space coverage translates into empirical gains on a variety of sparse reward navigation tasks (Section~\ref{sec:res:deep_RL}) and continuous control environments (Section~\ref{sec:res:continuous_control}). These results demonstrate the effectiveness of using the marginal state entropy as a regularizer that induces exploration. 


\subsection{Does Regularizing Entropy of the State Distribution Make Sense?}
\label{sec:res:hazan}

In this section, we expand the results from \citet{kakade_entropy} to validate the hypothesis that regularizing policy gradients with the entropy of state distribution, $\ent(d_\pi)$, can be useful for improved performance.

\textbf{Experimental Setup:} We use Frozen Lake \citep{brockman2016openai} where the entire state space is enumerable to obtain reward and transition dynamics. In this environment, we can compute \textit{exact policy gradient} with the state distribution entropy, $\ent(d_\pi)$, and policy entropy $\ent(\pi)$, weighted by regularization parameters $\lambda_s$ and $\lambda_\pi$ respectively.

\begin{figure}[t]
\begin{center}
    \begin{subfigure}[b]{0.45\columnwidth}
    \centering
        \includegraphics[width=1.\columnwidth]{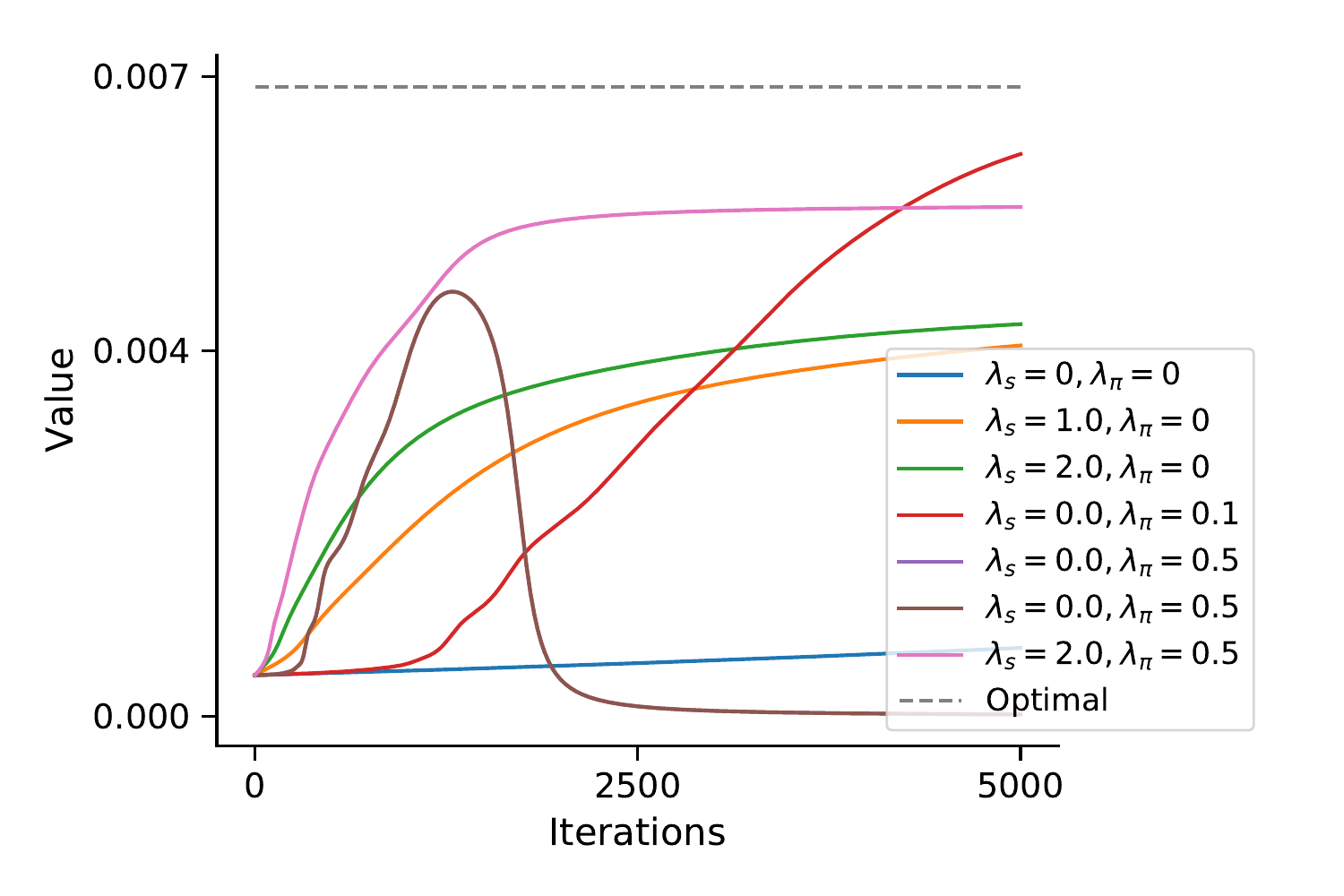}
        \caption{\label{fig:exact_pg}}
    \end{subfigure}
    \begin{subfigure}[b]{0.45\columnwidth}
    \centering
        \includegraphics[width=1.\columnwidth]{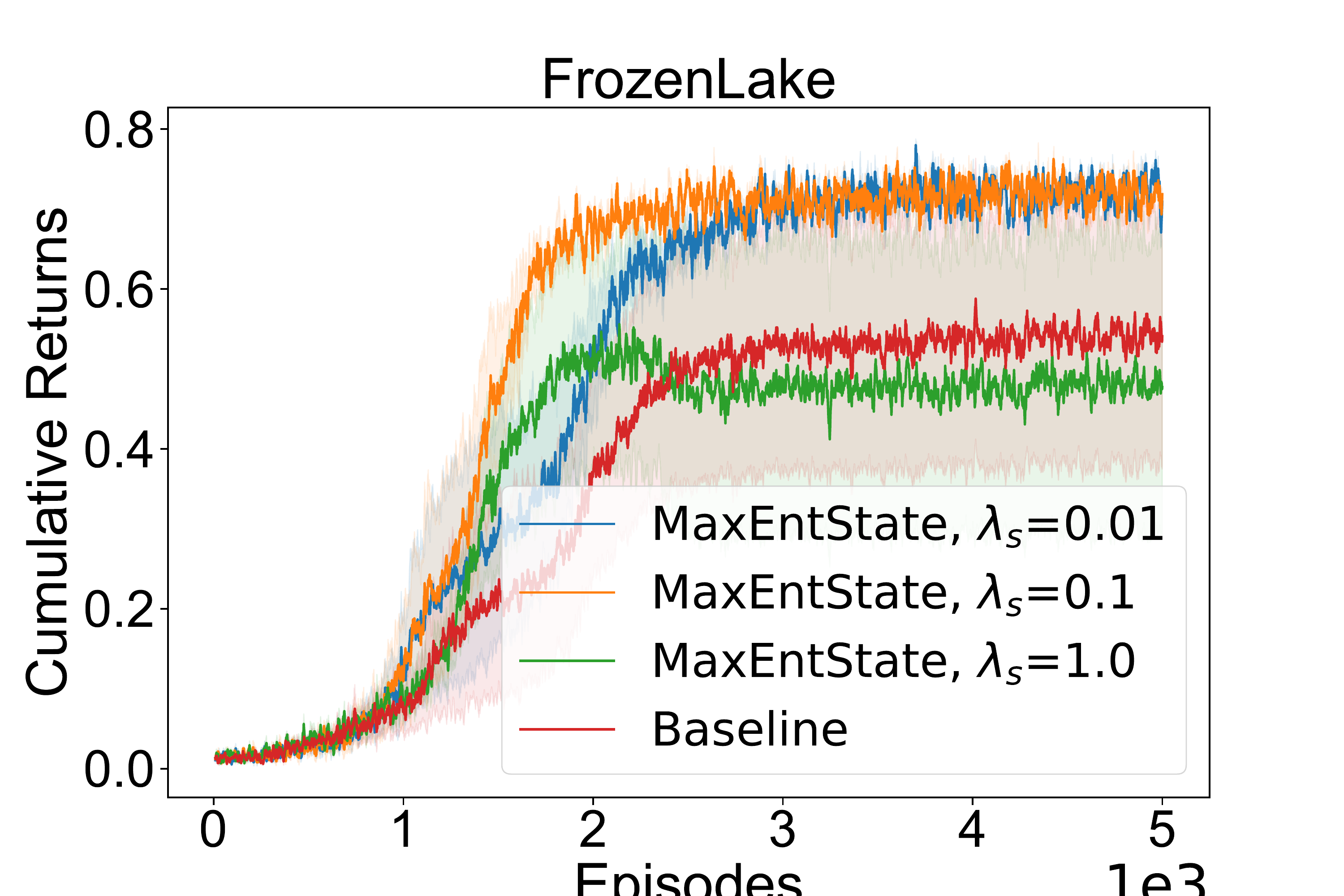}
        \caption{}
        \label{fig:tabular_ac}
    \end{subfigure}
    \caption{\small \textbf{Proof-of-concept experiments indicating the utility of maximizing state distribution entropy and our corresponding approximation, marginal state distribution entropy, on FrozenLake:} (a) A combination of state distribution entropy, $\ent(d_\pi)$, with $\lambda_s>0$, and policy entropy, $\ent(\pi)$, with $\lambda_\pi>0$, performs best in an exact policy gradient algorithm (b) Adding the marginal state entropy regularization $\ent (p_\pi)$ improves upon the performance compared with the policy entropy baseline, $\ent (\pi)$. See Figure~\ref{fig:frozen_lake_experiments_full}~and~\ref{fig:frozen_lake_experiments_full_interaction} for a closer analysis of $\lambda_\pi$ and $\lambda_s.$ 
    \label{fig:frozen_lake_experiments}
    }
\end{center}    
\vspace{-2em}
\end{figure}

\textbf{Result:} We find that using state distribution entropy in addition to policy entropy can lead to faster learning compared to them in isolation (Figure~\ref{fig:exact_pg}). Furthermore, we found that while policy entropy performed quite well (Figure~\ref{fig:exact_pg}~and~\ref{fig:frozen_lake_policy_ent} in Appendix) with respect to the solution found, it required a decay on $\lambda_\pi$ which was was difficult to find (Figure~\ref{fig:frozen_lake_pol_ent_w_decay}). In contrast, state distribution entropy had a stable performance for a wide range of hyperparameters (Figure~\ref{fig:frozen_lake_state_ent}~and~\ref{fig:frozen_lake_state_ent_w_decay}, addition) despite reaching a suboptimal solution. This is not surprising as the optimal solution will not be the one that maximizes state space coverage. Additionally, state distribution entropy regularization has a stabilizing interactive effect on policy entropy regularization (Figure~\ref{fig:frozen_lake_experiments_full_interaction} as shown in appendix).

\subsection{State Space coverage in Complex GridWorlds}
\label{sec:res:state_space_coverage}
The maximization of state distribution entropy should lead to a more uniform coverage of the state space. In the following section, we empirically verify increased state space coverage compared to the baseline of maximizing policy entropy in Grid World domains.

\textbf{Experimental Setup:} To confirm our intuitions about state space coverage, we use three environments: (1) Pachinko where a simple grid world is periodically dotted with impassable walls; (2) Double-slit where the agent starts at the extreme bottom left and must navigate three rooms with small doors to reach a goal at the extreme top right; and (3) Four rooms \citep{Schaul}. The agents for (1) and (2) are trained with \textsc{Reinforce} \citep{reinforce} while the agent for (3) is trained using a simple actor-critic method with GAE \citep{konda2000actor,schulman2015high}.

\begin{figure}[b]
\begin{center}
    \begin{subfigure}[b]{0.23\columnwidth}
    \centering
        \includegraphics[width=1.\columnwidth]{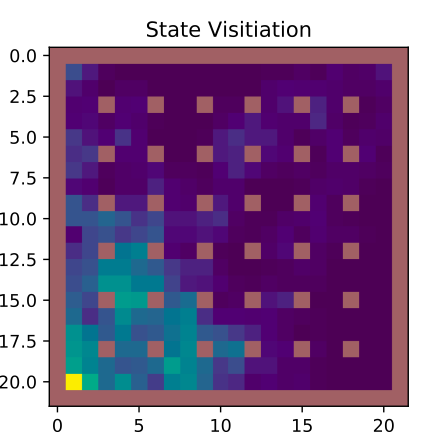}
        \caption{MaxEntPolicy}
        \label{fig:pachinko_maxentpolicy}
    \end{subfigure}
    \begin{subfigure}[b]{0.23\columnwidth}
    \centering
        \includegraphics[width=1.\columnwidth]{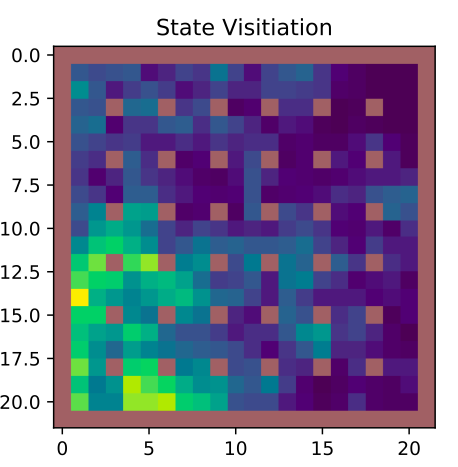}
        \caption{MaxEntState}
        \label{fig:pachinko_maxentstate}
    \end{subfigure}
    \begin{subfigure}[b]{0.23\columnwidth}
    \centering
        \includegraphics[width=1.\columnwidth]{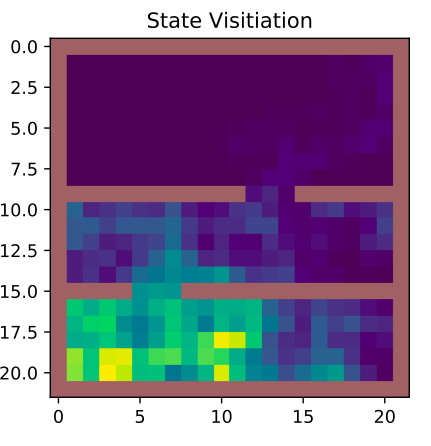}
        \caption{MaxEntPolicy}
        \label{fig:grid_maxentpolicy}
    \end{subfigure}
    \begin{subfigure}[b]{0.23\columnwidth}
    \centering
        \includegraphics[width=1.\columnwidth]{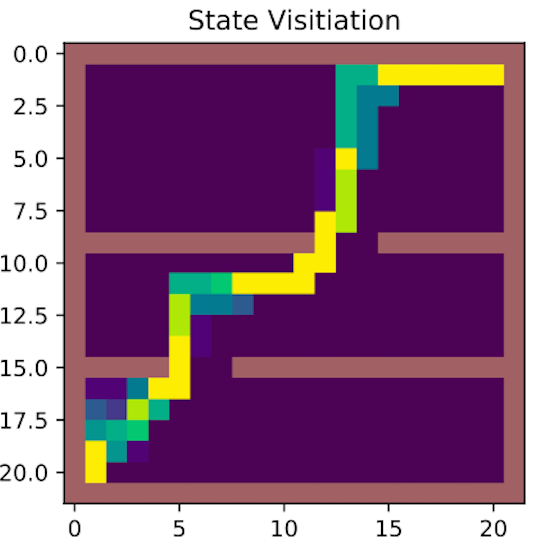}
        \caption{MaxEntState}
        \label{fig:grid_maxentstate}
    \end{subfigure}
    \caption{\small \textbf{Improved state space coverage on a complex gridworld leads to better policies}: A Pachinko world demonstrates that (a) using MaxEntPolicy regularization produces a policy that centers around the starting position while (b) MaxEntState leads to a more dispersed policy with wider state space coverage. In a double-slit environment, the agent must navigate three rooms to reach the goal. (c) MaxEntPolicy results in a random walk policy that leaves the first room but never makes it into the third whereas (d) MaxEntPolicy successfully reaches the goal in the third room. We highlight the fact that it would never be able to reach this room without leaving the second room. Light colored regions represent areas where a trained policy spent time in a grid.\label{fig:complexgrids}}
\end{center}    
\vspace{-1em}
\end{figure}

\textbf{Results:} Our first result provides evidence for the validity of our approximation of the state distribution. When using MaxEntPolicy the agent does not move far from the starting position (Figure~\ref{fig:pachinko_maxentpolicy}) indicating that a more random policy will not successfully navigate beyond a few walls. In contrast, MaxEntState (Figure~\ref{fig:pachinko_maxentstate}) is able to reach a larger portion of the state space providing empirical validation that the technique encourages the policy to navigate into a wider area of the world. In the double-slit environment, MaxEntPolicy finds a mostly random policy and only ends up visiting a small region of the grid, barely making it to the other rooms (Figure~\ref{fig:grid_maxentpolicy}). In contrast, MaxEntState successfully finds a policy that reaches the goal state in the last room (Figure~\ref{fig:grid_maxentstate}).

\begin{figure}[t]
\begin{center}
    \begin{subfigure}[b]{0.28\columnwidth}
    \centering
        \includegraphics[width=1.\columnwidth]{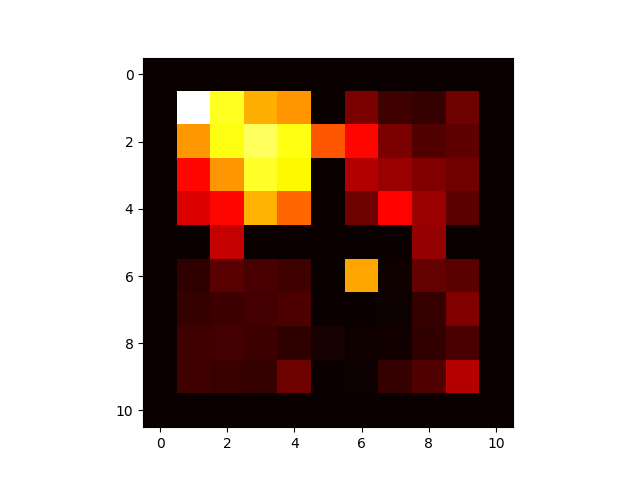}
        \caption{MaxEntPolicy\label{fig:FR_MaxEntPolicy}}
    \end{subfigure}
    \begin{subfigure}[b]{0.28\columnwidth}
    \centering
        \includegraphics[width=1.\columnwidth]{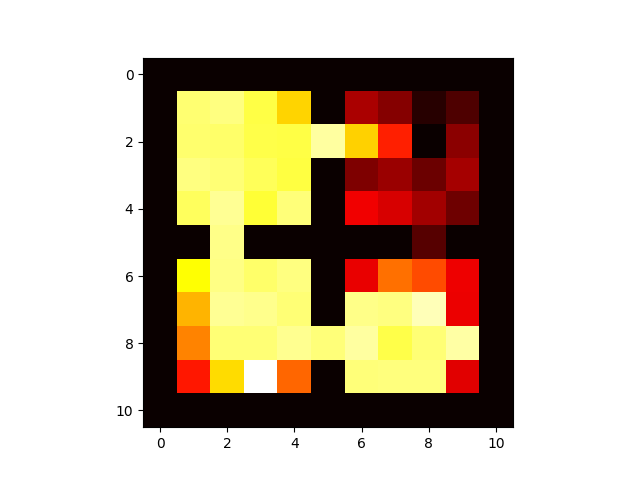}
        \caption{MaxEntState\label{fig:FR_MaxEntState}}
    \end{subfigure}
    \quad
    \begin{subfigure}[b]{0.4\columnwidth}
    \centering
        \includegraphics[width=1.\columnwidth]{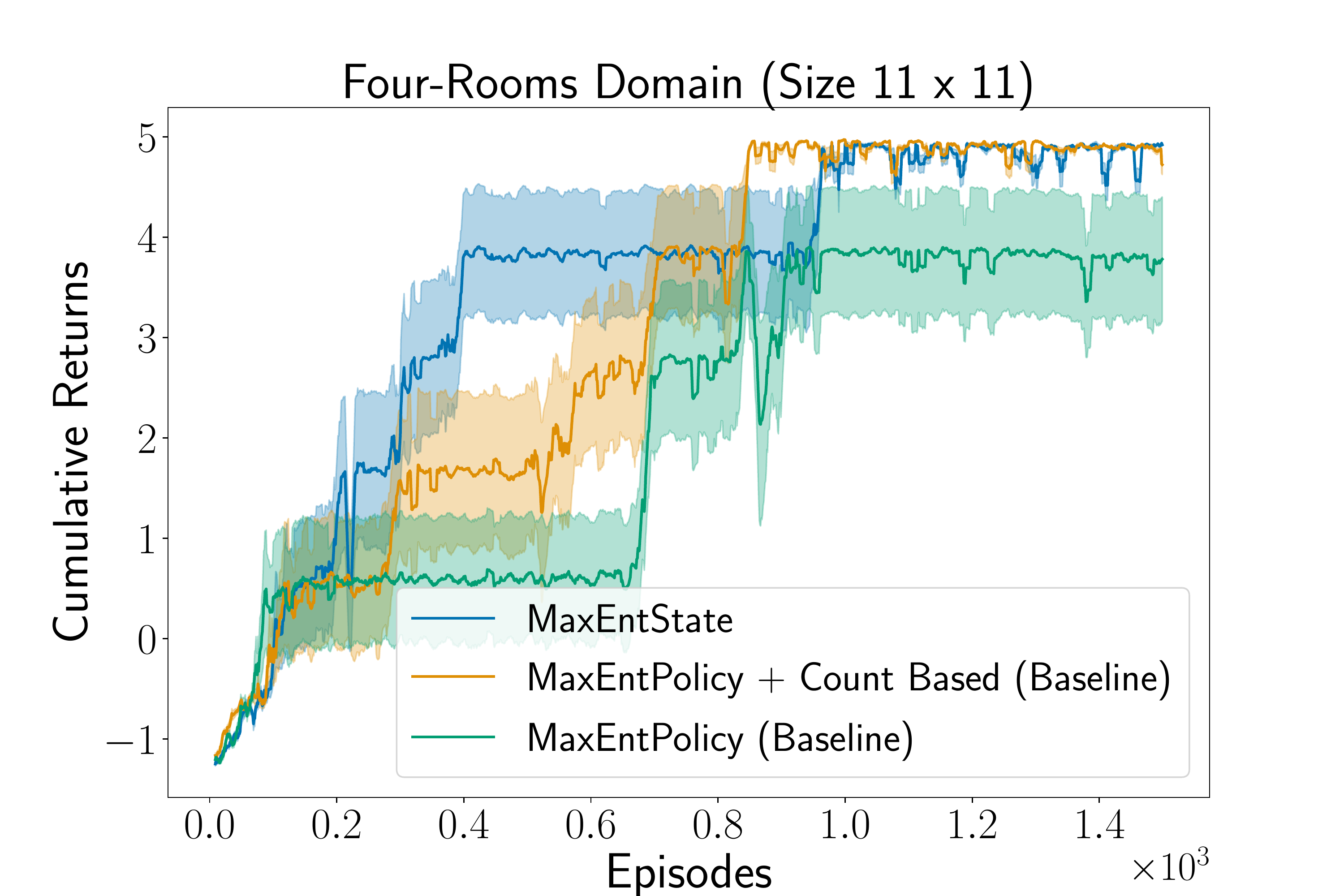}
        \caption{Learning Curves\label{fig:FR_Reward}}
    \end{subfigure}
    \caption{\small \textbf{MaxEnt has improved exploration and sample efficiency in the Four Rooms Domain.} Policies trained with MaxEntState in (b) visit a larger region of the gird compared to MaxEntPolicy in (a). Cumulative returns plot in (c) comparing MaxEntState regularization with baselines (count based exploration and MaxEntPolicy) show improved sample efficiency.\label{fig:fourrooms}}
\end{center}   
\vspace{-2em}
\end{figure}

To quantify if this improved exploration translates to improved learning speed, we measure both the qualitative heatmaps as well as the return on the four room domain during learning. Consistent with the other environments, the final policy learned using MaxEntState (Figure~\ref{fig:FR_MaxEntState}) has an increased state space coverage compared to a policy learned with MaxEntPolicy alone (Figure~\ref{fig:FR_MaxEntPolicy}). These qualitative results translate into quantitative improvements in learning performance.


\subsection{MaxEntState Improves Performance on Partially Observed Sparse Reward Tasks}
\label{sec:res:deep_RL}


The results from Section~\ref{sec:res:state_space_coverage}, and in particular Figure~\ref{fig:complexgrids}, suggest that agents which maximize state distribution entropy do indeed maximize state space coverage. We expect that these qualitative traits provide empirical performance improvements in complicated sparse-reward and partially observable domains. In this section, we quantitatively measure performance improves in deep RL by considering a variety of partially observed 2D and 3D tasks.

\textbf{Experimental Setup:} To measure quantitatively the impact of using MaxEntState, we consider a range of partially observable sparse reward 2D environments generated with MiniGrid \citep{gym_minigrid} and 3D environments generated with MiniWorld\footnote{These environments are alternative to the VizDoom or DeepMind Lab environments.} \citep{gym_miniworld}. In MiniGrid, the environments are setup as 2D grids and the agent only receives a small area as an observation (Figure~\ref{fig:minigrid_env_plots}). In MiniWorld, the agent receives a high dimensional image of a 3D space and then must learn to navigate a complex maze consisting of rooms, doors and hallways. Both environments are challenging to solve in practice because a positive reward is only given at the end of a successful episode. For both the MiniGrid and Miniworld environments, we use both A2C \citep{mhiha2c} and PPO \citep{PPO} policy gradient algorithms. 


\begin{figure}[hb]
\begin{center}
    \begin{subfigure}[b]{0.31\columnwidth}
    \centering
        \includegraphics[width=1.\columnwidth]{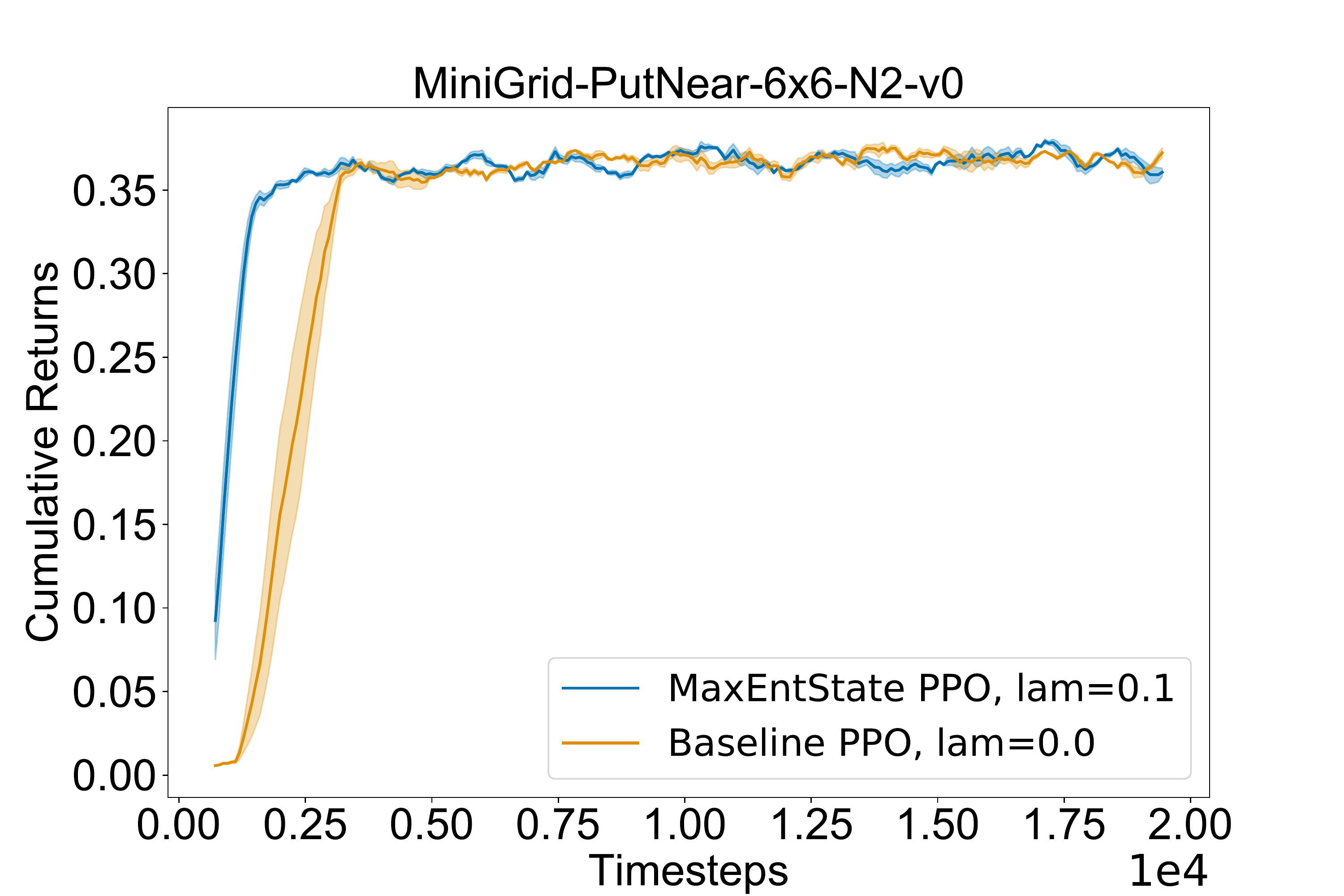}
        \caption{}
        \label{fig:putnear8}
    \end{subfigure}
    \quad
    \begin{subfigure}[b]{0.31\columnwidth}
    \centering
        \includegraphics[width=1.\columnwidth]{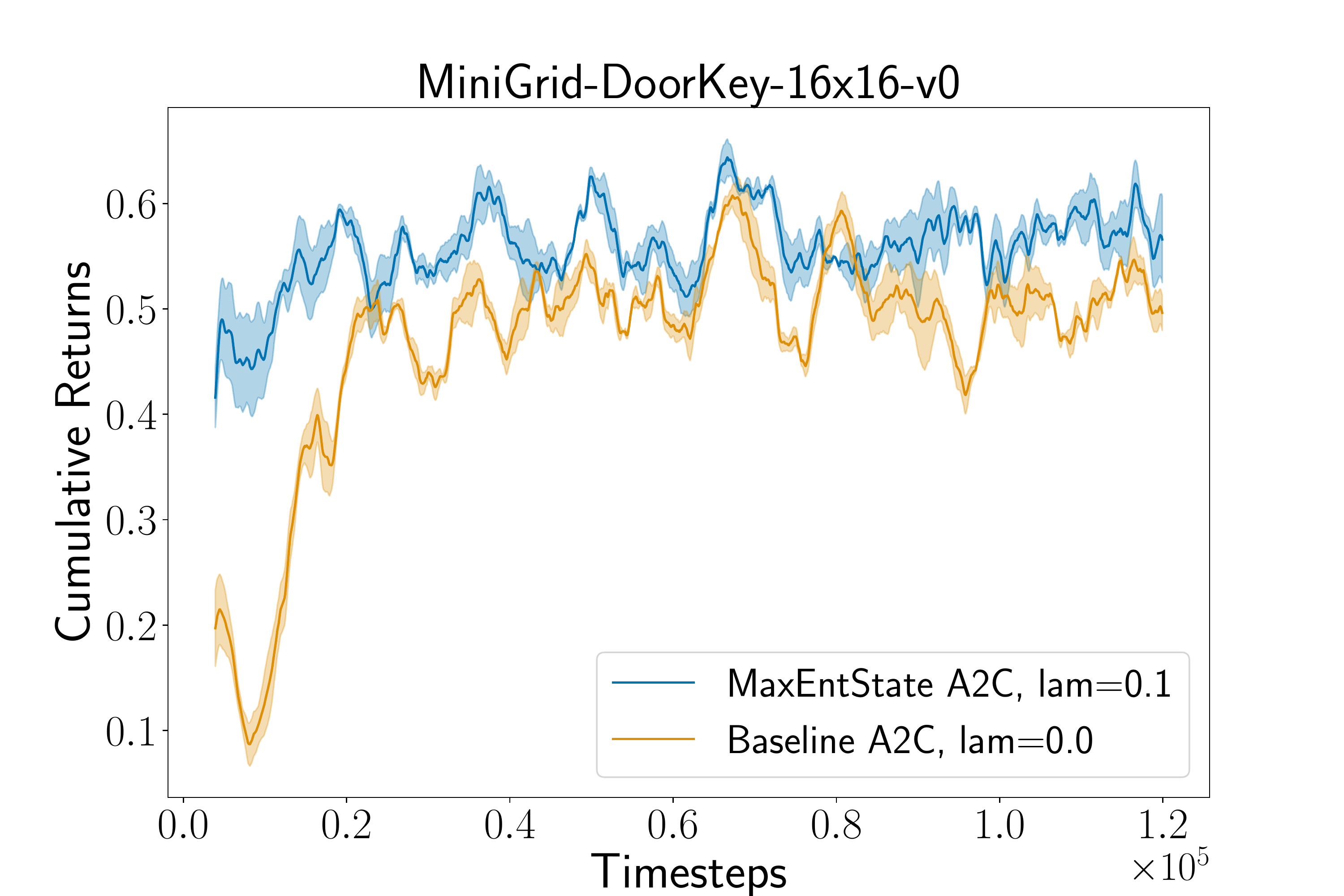}
        \caption{}
        \label{fig:doorkey16}
    \end{subfigure}
    \quad
    \begin{subfigure}[b]{0.31\columnwidth}
    \centering
        \includegraphics[width=1.\columnwidth]{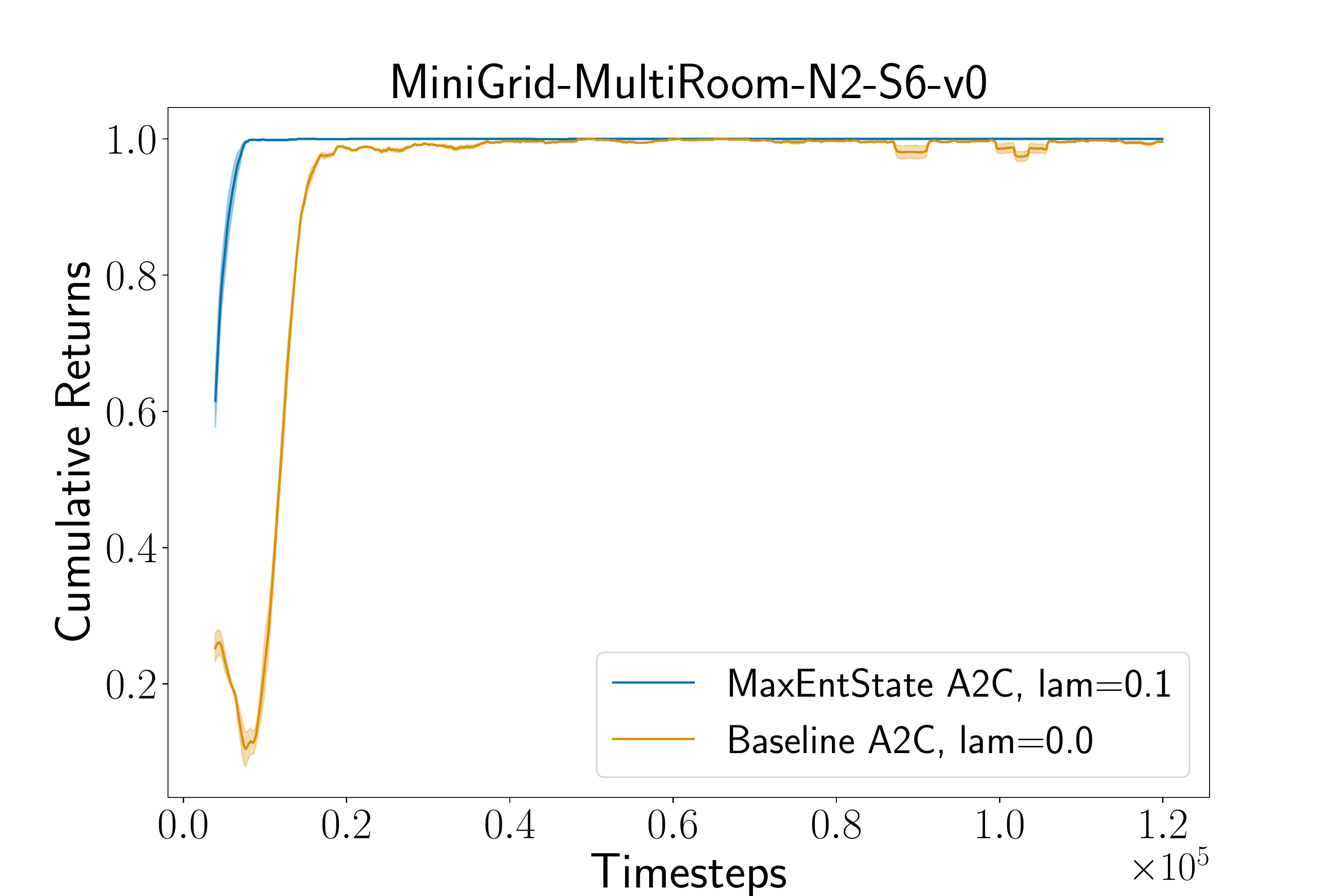}
        \caption{}
        \label{fig:multiroom}
    \end{subfigure}
    \caption{\small \textbf{MaxEntState improves learning speed on challenging 2D partially observable environments.} MaxEntState (blue), the agent quickly is quickly able to find the rewards in (a) \textit{PutNearSXNY} (b) \textit{DoorKeySXNY} and (c) \textit{MultiRoomSXNY} compared to using MaxEntPolicy alone (Baseline, orange). NX denotes the sizes of the rooms and NY denotes the number of rooms. PutNear involves placing objects near a target while DoorKey involves finding a key before navigating into a room with the goal.\label{fig:minigird_results_a2c}}
\end{center}    
\end{figure}

\begin{figure}[ht]
\begin{center}
    \begin{subfigure}[b]{0.3\columnwidth}
    \centering
        \includegraphics[width=1.\columnwidth]{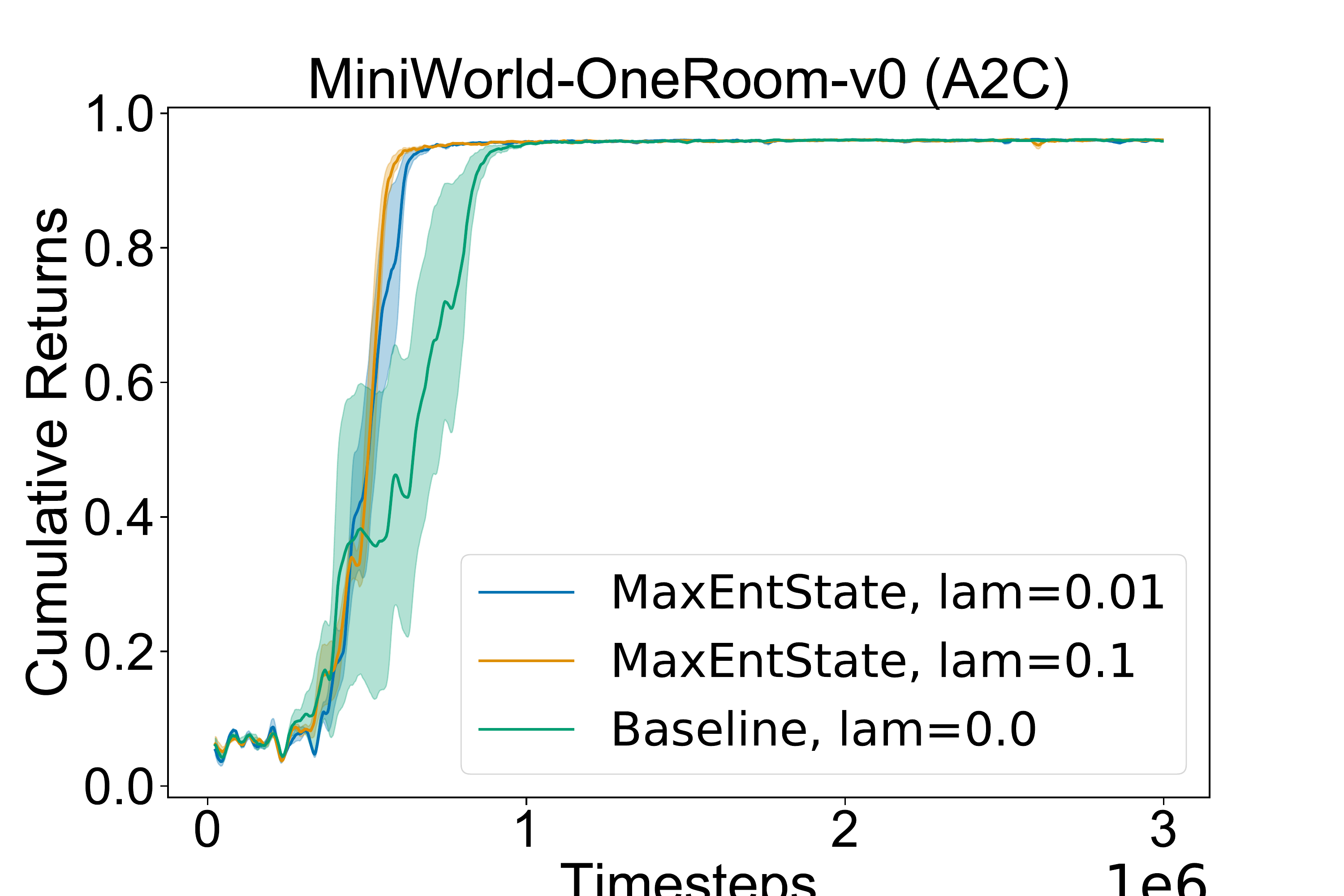}
        \caption{\label{miniworld_oneroom}}
    \end{subfigure}
    \quad
    \begin{subfigure}[b]{0.3\columnwidth}
    \centering
        \includegraphics[width=1.\columnwidth]{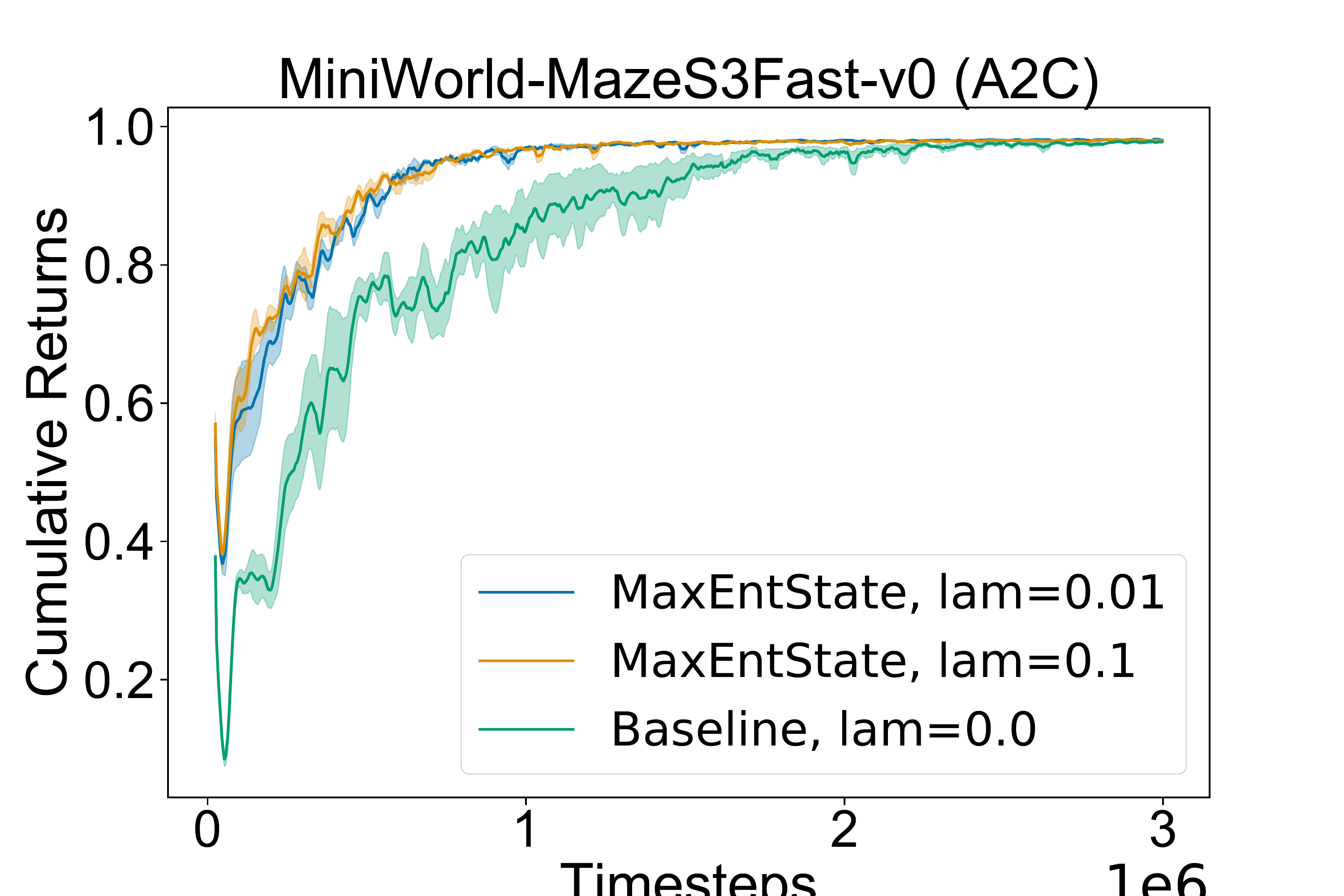}
        \caption{\label{miniworld_mazes3fast}}
    \end{subfigure}
    \quad
    \begin{subfigure}[b]{0.3\columnwidth}
    \centering
        \includegraphics[width=1.\columnwidth]{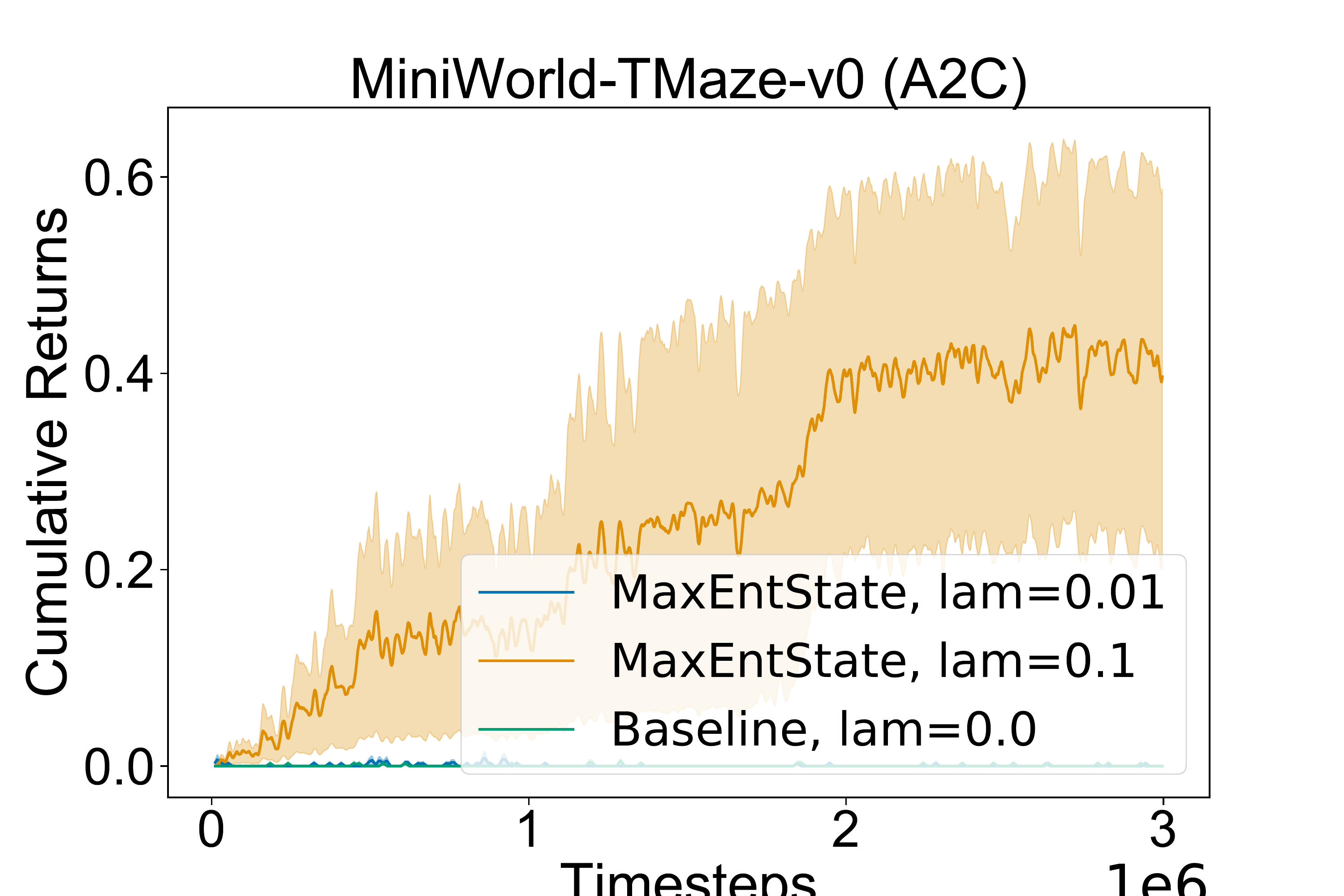}
        \caption{\label{miniworld_tmaze}}
    \end{subfigure}\\
    \quad
    \begin{subfigure}[b]{0.3\columnwidth}
    \centering
        \includegraphics[width=1.\columnwidth]{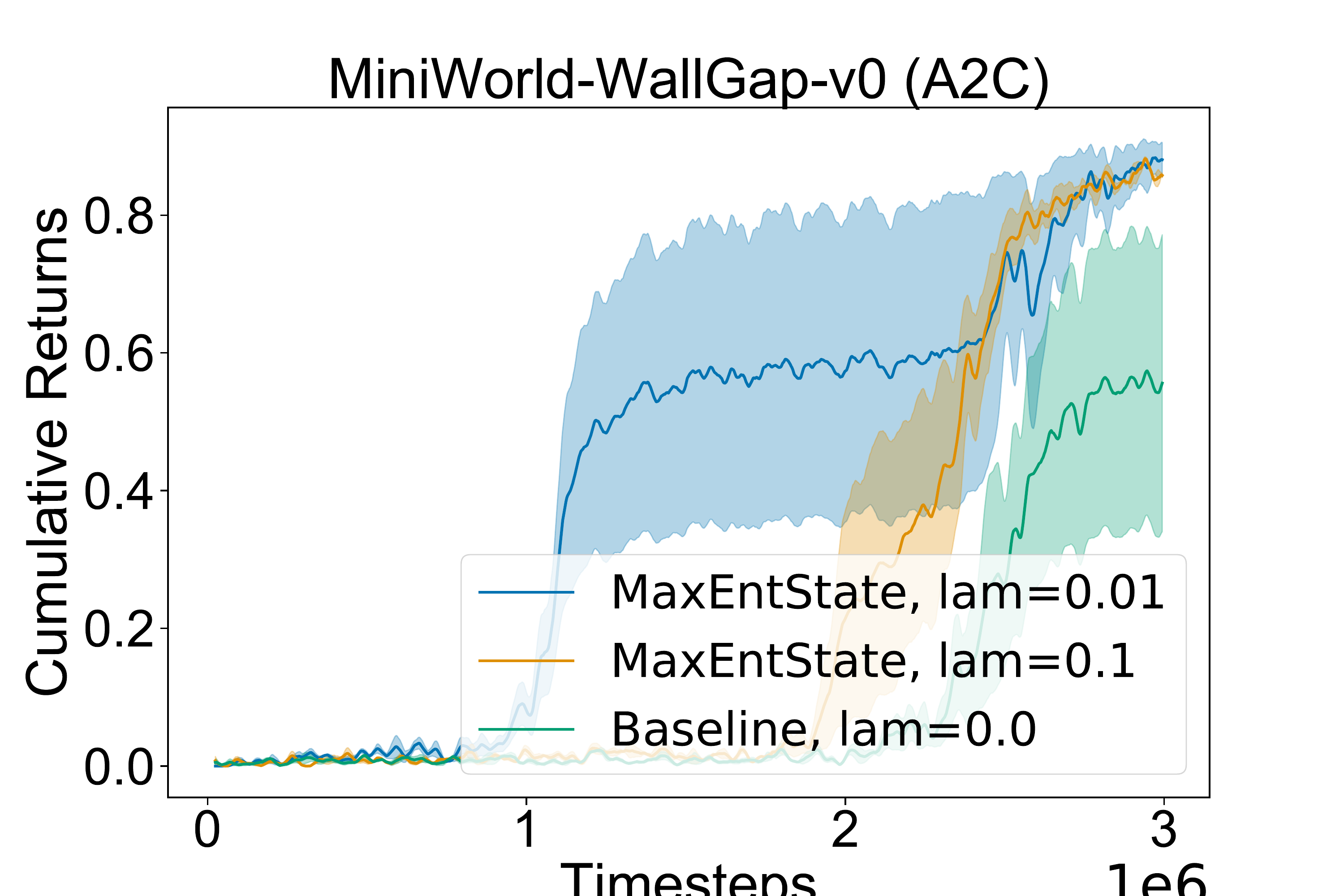}
        \caption{\label{miniworld_wallgap}}
    \end{subfigure}  
    \quad
    \begin{subfigure}[b]{0.3\columnwidth}
    \centering
        \includegraphics[width=1.\columnwidth]{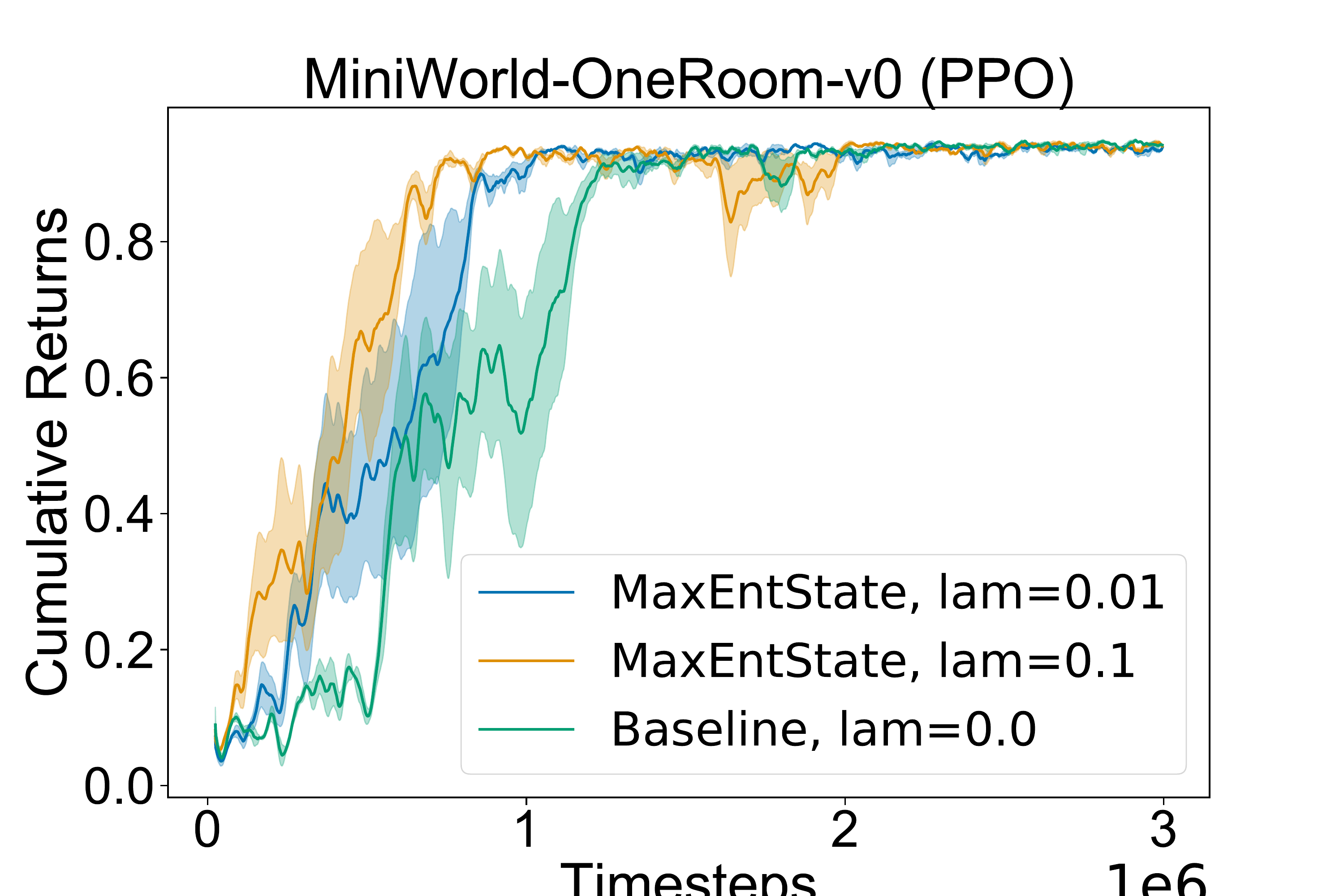}
        \caption{\label{miniworld_oneroom_ppo}}
    \end{subfigure}
    \quad
    \begin{subfigure}[b]{0.3\columnwidth}
    \centering
        \includegraphics[width=1.\columnwidth]{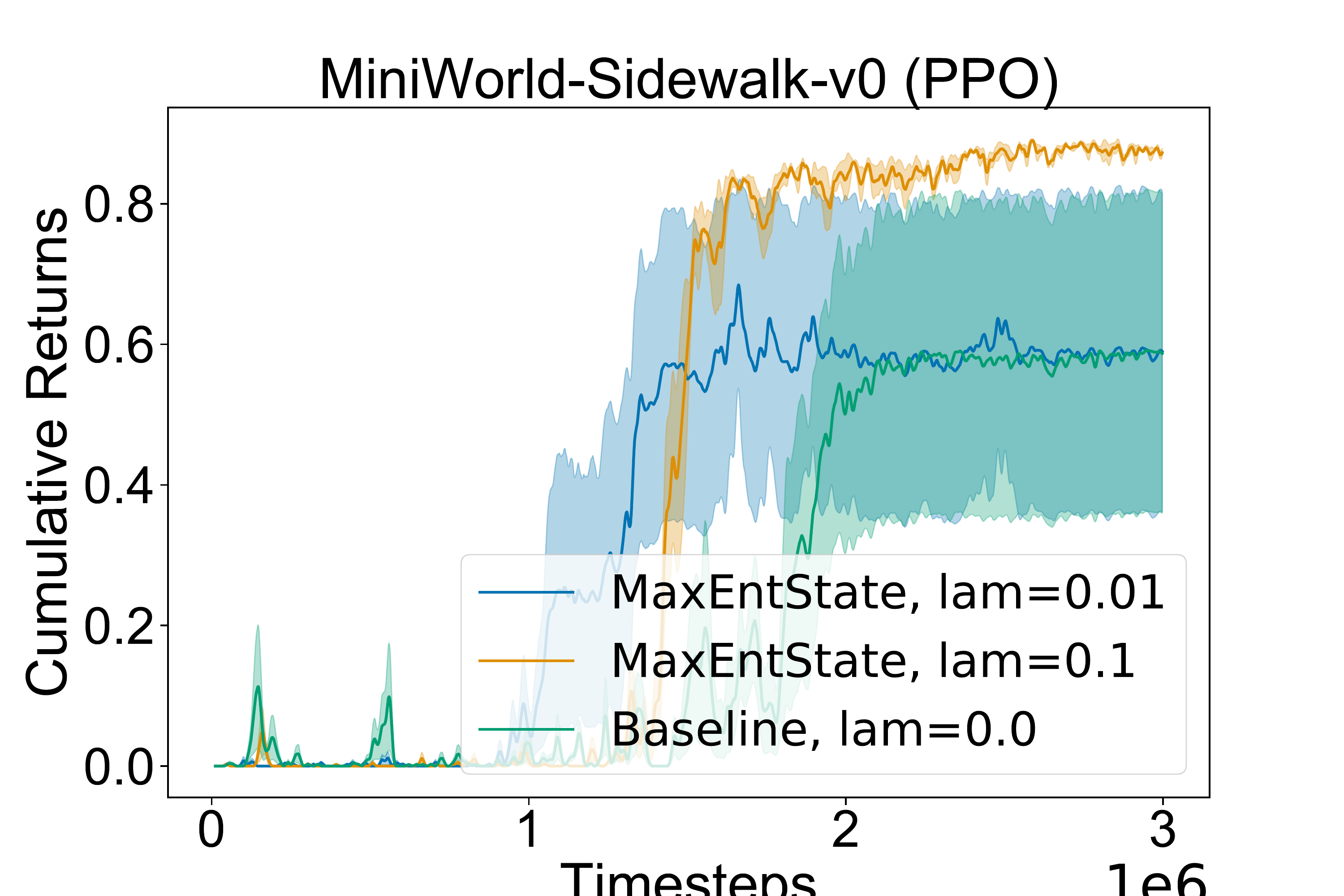}
        \caption{\label{miniworld_sidewalk_ppo}}
    \end{subfigure}
    
    \caption{\small \textbf{MaxEntState provides improved exploration to solve (partially) difficult sparse reward environments}. MiniWorld Envoronments with A2C, with different $\lambda$ weightings for the state entropy regularization. The baseline (green) is a standard A2C with policy entropy regularization ($\lambda_\pi=0.1, \lambda_s=0.0$). Additional results using PPO can be seen in Figure~\ref{fig:miniworld_ppo} in Appendix.\label{fig:miniworld}}
\end{center}    
\end{figure}

\textbf{Results:} MaxEntState improves learning speed on all MiniGrid environments tested (Figure~\ref{fig:minigird_results_a2c}). In particular, the agent consistently finds the goal within the first 1000-2000 timesteps of experience. In the more complicated 3D tasks, MaxEntState entropy regularization consistently performs better compared to MaxEntPolicy baseline (Figure~\ref{fig:miniworld}~and~\ref{fig:miniworld_ppo}). We were able to find some environments where only policies trained using MaxEntState were able to learn to solve the task (Figures~\ref{miniworld_tmaze}, \ref{miniworld_wallgap_ppo},~and~\ref{miniworld_tmaze_ppo}). These environments are quite difficult for MaxEntPolicy baseline to solve with random exploration, since they contain 3D observation space, are partially observable and sparse reward. In these tasks, MaxEntPolicy typically fails, as exploration of the state space plays a key role in these multiple room navigation tasks.

\subsection{MaxEntState on Continuous Control}
\label{sec:res:continuous_control}
Finally, we compare the effectiveness of marginal state entropy regularization for common continuous control tasks. In control environments, exploration of the state space often plays a key role for sample efficiency in solving these tasks.

\textbf{Experimental Setup:} We use the popular MuJoCo simulator for continuous control \citep{conf/iros/TodorovET12} with environments from OpenAI Gym \citep{brockman2016openai}. We us the Soft Actor-Critic (SAC) \citep{sac} and Deterministic Policy Gradient (DDPG) \citep{DDPG} frameworks. We use open-source tuned implementations of DDPG and SAC, as these algorithms have high variance and often unstable to use in practice \citep{henderson}. In DDPG, we use MaxEntState as an additional state entropy regularizer, and compare with baseline DDPG only. In SAC, we use MaxEntState regularizer additionally with the MaxEnt framework with $\lambda_{\pi}=0.1$.

\textbf{Results:} We find that using MaxEntState on continuous control problem provide get marginal improvements using SAC (Figure~\ref{fig:SAC}). These improvements are much larger in DDPG (Figure \ref{fig:DDPG} in Appendix), since baseline DDPG uses deterministic policies only, and exploration is achieved with random noise in action space only.

\begin{figure}[h]
\begin{center}
    \begin{subfigure}[b]{0.3\columnwidth}
    \centering
        \includegraphics[width=1.\columnwidth]{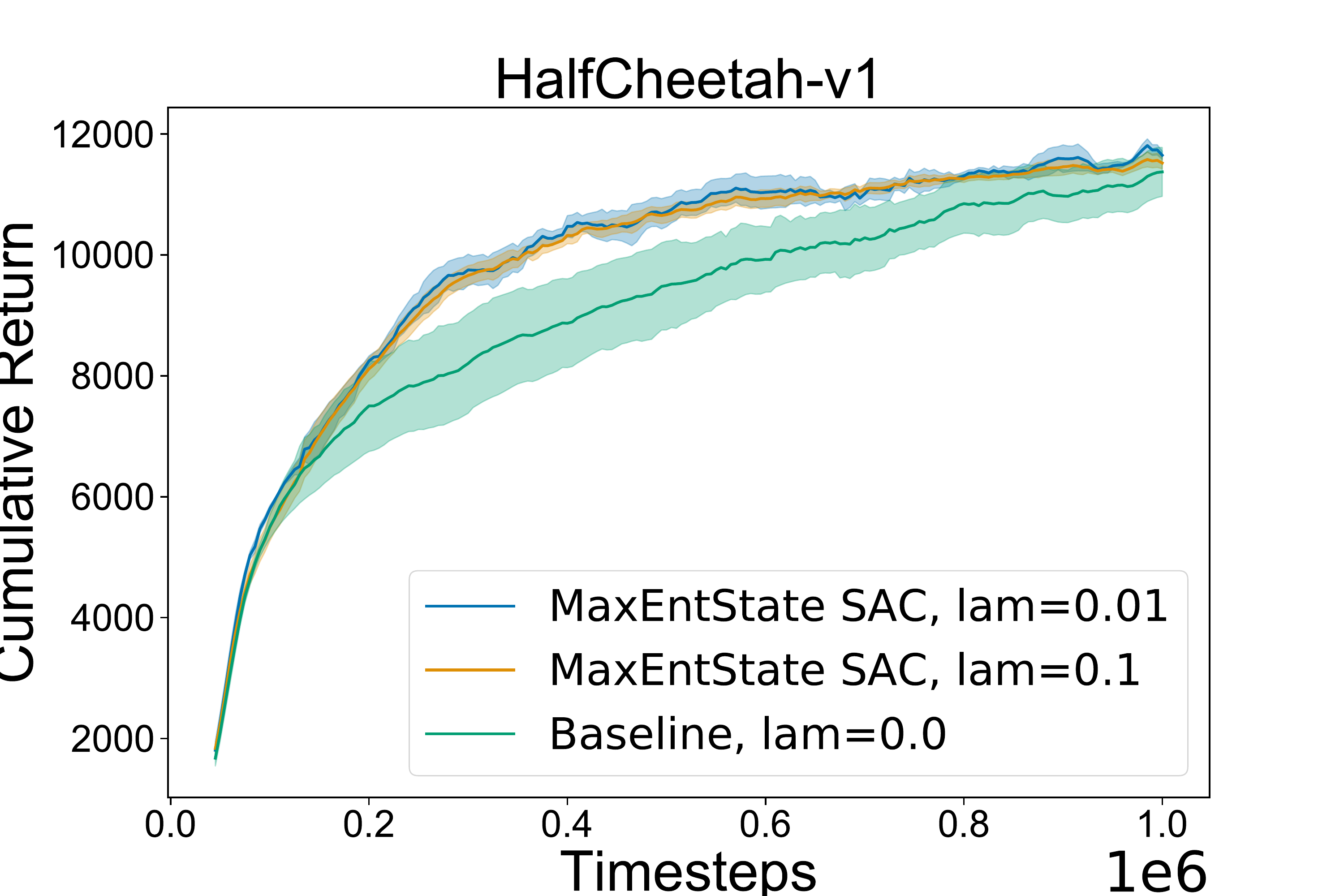}
        \caption{\label{sac_halfcheetah}}
    \end{subfigure}
    \begin{subfigure}[b]{0.3\columnwidth}
    \centering
        \includegraphics[width=1.\columnwidth]{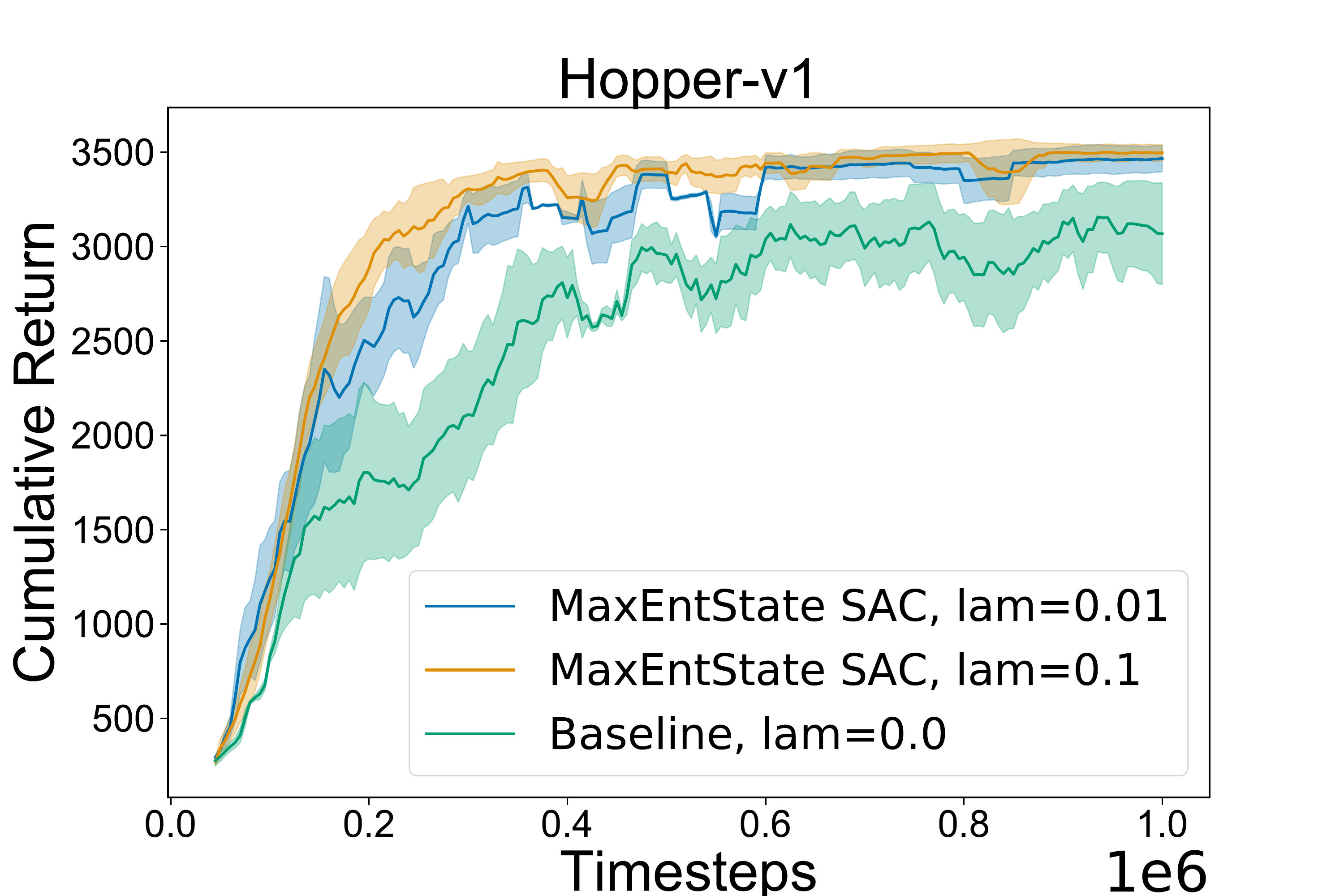}
        \caption{\label{sac_hopper}}
    \end{subfigure}
    \quad
    \begin{subfigure}[b]{0.3\columnwidth}
    \centering
        \includegraphics[width=1.\columnwidth]{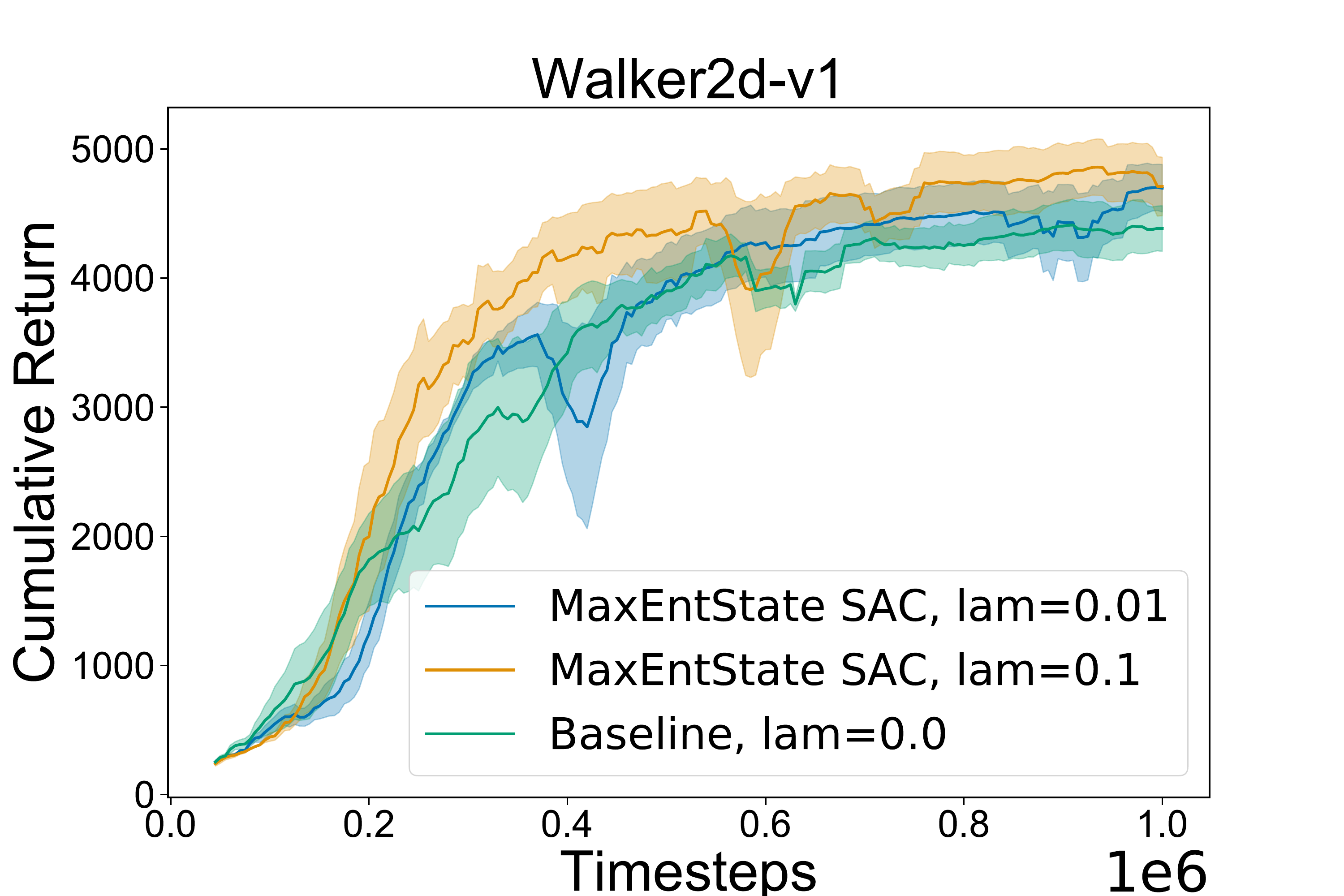}
        \caption{\label{sac_walker}}
    \end{subfigure}
    \caption{\small \textbf{Slight improvements in control domains with state space exploration}. Continuous Control Benchmarks on Soft Actor-Critic. We compare with SAC here only, since it uses MaxEnt framework for exploration and improved stability, and we compare with and without state entropy regularization. We achieve more significant improvements with DDPG as shown in Figure~\ref{fig:DDPG}.  \label{fig:SAC}}
\end{center}    
\end{figure}

\section{Related Work}

MaxEntState joints a family of methods that does reward shaping \citep{NgRewardShaping}, by augmenting external rewards, $r$, from the environment with an additional reward signal, $b$. While the external reward, $r$, is set by the environment, we are free to pick the internal reward, $b$: Some examples of $b$ include using a hash-map based count of the states visited \citep{tang2017exploration}, the norm of the successor feature \citep{machado2018count} and feature space prediction error \citep{pathak2017curiosity}. Particularly related are count-based models for exploration. Following pseudo-count based models \citep{Bellmare_Count}, the empirical distribution, $\hat{p}(s)$ can be measured by the number of occurrences of a state $s$ in the sequence, $s_{1:n}$. This pseudo-count $\hat{p}(s)$ can also be defined in terms of a density model, $\rho$ \citep{bellemare_density}. The entropy of the induced state distribution, $\ent(\hat{p}(s))$ or $\ent(\rho(s))$, can thus be measured and provided as an intrinsic motivation similar to MaxEntState. Unlike our work, no gradient of this term exists with respect to the policy parameters.



Our method deviates from the usual intrinsic motivation algorithms as MaxEntState also serves as a regularizer. Specifically, there exists a gradient signal given by the bonus, $\nabla_\theta b$, with respect to the policy parameters, $\theta$ \citep{schulman2017equivalence}. Regularizers are often used to extract specific kinds of behaviour to exploit structure of certain tasks. For example, mutual information regularizes can limit the dependence of the policy actions on the environment goals encouraging more generalizable behaviour \citep{infobot,galashov2019information}. Additionally, certain maximum entropy regularizers can encourage a diversity amongst policies \citep{bachman2018vfunc,eysenbach2019diversity}, enhance state space coverage \citep{kakade_entropy} and diversity of goals \citep{pong2019skew}.

\section{Discussions}
In this work, we have shown that regularizing with maximum state entropy can perform better than using maximum entropy policy by itself. 

\textbf{When and why does maximizing the entropy of the marginal state distribution matter?} 
In this work, we provide a regularized policy gradient which directly influences state space coverage for exploration. Our results show, through visualizations of the state space in gridworlds, that the marginal state space regularization does indeed improve the state space coverage (Figure~\ref{fig:pachinko_maxentstate}). However, we do not expect a state distribution regularization scheme to work everywhere: Policies that maximize state space coverage are effectively sub-optimal (Figure~\ref{fig:exact_pg}). This can be problematic when there is a dense reward signal and our agent is being incentivized to simultaneously follow the reward as well as maximize state space coverage. Despite this intuition, we have shown that there is a small benefit in dense reward continuous control tasks as well.

In contrast, we do expect that the gains from this state space coverage translate to quicker learning in environments with sparse or no reward structure: In such environments, a \emph{good} random exploration strategy is important. Policy entropy and state distribution entropy both induce two very different kinds of random exploration. While a mechanism that increases the randomness of policies will do a random walk in space (Figure~\ref{fig:pachinko_maxentpolicy}~and~\ref{fig:FR_MaxEntPolicy}), a mechanism that visits novel states is likely to be more successful in discovering an eventual reward. Indeed, we found this to be true for the proposed algorithm in the sparse reward environments we have tested here. We expect that, decaying the weighting term, $\lambda_\pi$, should eventually recover the true objective we care about, the discounted sum of rewards.

\textbf{Limitations and Future Work:} One major limitation of our approach is that we use the marginal, per-step state distribution for entropy,  instead of maximizing the entropy of the \textit{stationary} or \textit{discounted} state distributions. Maximizing the stationary state distribution can directly influence coverage as  stationary would imply \textit{exploration independent of time}, whereas discounted state distribution based on future occupancy measures would be more reliable means of exploration in episodic settings. In practice, the stationary or discounted state distribution can be directly estimated by learning an explicit density estimator that is directly influenced by the policy parameters.

\textbf{Conclusion:} This work proposes an entropy regularization technique in policy optimization based on maximizing the marginal state distribution $\ent (p_{\pi})$, which can be used as an approximation to $\ent (d_{\pi})$. We provide a  simple mechanism that can be used on top of any existing policy gradient algorithm to directly influence state space coverage. We showed that this scheme can learn policies that induce a larger state space coverage which can be an effective exploration objective. Finally, this regularization improved performance on a range of sparse reward environments. In closing, we believe our work provides a step towards extending entropy regularization that can directly influence state space coverage, which can perhaps tackle fundamental problems of exploration in reinforcement learning.


\subsection*{Acknowledgements}
We thank Anirudh Goyal for preliminary discussions regarding this work. We are grateful to Emmanuel Bengio, Pascal Lamblin, John Martin and Pablo Samuel Castro for comments on drafts of this manuscript. This research was enabled in part by support provided by \href{CalculQuebec}{http://www.calculquebec.ca/en/} and \href{Compute Canada}{www.computecanada.ca}.

\bibliography{neurips_2019}

\begin{thebibliography}{32}
\providecommand{\natexlab}[1]{#1}
\providecommand{\url}[1]{\texttt{#1}}
\expandafter\ifx\csname urlstyle\endcsname\relax
  \providecommand{\doi}[1]{doi: #1}\else
  \providecommand{\doi}{doi: \begingroup \urlstyle{rm}\Url}\fi

\bibitem[Mnih et~al.(2016)Mnih, Badia, Mirza, Graves, Lillicrap, Harley,
  Silver, and Kavukcuoglu]{mhiha2c}
Volodymyr Mnih, Adri{\`{a}}~Puigdom{\`{e}}nech Badia, Mehdi Mirza, Alex Graves,
  Timothy~P. Lillicrap, Tim Harley, David Silver, and Koray Kavukcuoglu.
\newblock Asynchronous methods for deep reinforcement learning.
\newblock In \emph{Proceedings of the 33nd International Conference on Machine
  Learning, {ICML} 2016, New York City, NY, USA, June 19-24, 2016}, pages
  1928--1937, 2016.
\newblock URL \url{http://jmlr.org/proceedings/papers/v48/mniha16.html}.

\bibitem[Haarnoja et~al.(2018)Haarnoja, Zhou, Abbeel, and Levine]{sac}
Tuomas Haarnoja, Aurick Zhou, Pieter Abbeel, and Sergey Levine.
\newblock Soft actor-critic: Off-policy maximum entropy deep reinforcement
  learning with a stochastic actor.
\newblock In \emph{Proceedings of the 35th International Conference on Machine
  Learning, {ICML} 2018, Stockholmsm{\"{a}}ssan, Stockholm, Sweden, July 10-15,
  2018}, pages 1856--1865, 2018.
\newblock URL \url{http://proceedings.mlr.press/v80/haarnoja18b.html}.

\bibitem[Ahmed et~al.(2019)Ahmed, Roux, Norouzi, and
  Schuurmans]{understanding_entropy}
Zafarali Ahmed, Nicolas~Le Roux, Mohammad Norouzi, and Dale Schuurmans.
\newblock Understanding the impact of entropy on policy optimization.
\newblock \emph{International Conference on Machine Learning}, abs/1811.11214,
  2019.
\newblock URL \url{http://arxiv.org/abs/1811.11214}.

\bibitem[Hazan et~al.(2018)Hazan, Kakade, Singh, and Soest]{kakade_entropy}
Elad Hazan, Sham~M. Kakade, Karan Singh, and Abby~Van Soest.
\newblock Provably efficient maximum entropy exploration.
\newblock \emph{CoRR}, abs/1812.02690, 2018.
\newblock URL \url{http://arxiv.org/abs/1812.02690}.

\bibitem[Bellemare et~al.(2016)Bellemare, Srinivasan, Ostrovski, Schaul,
  Saxton, and Munos]{Bellmare_Count}
Marc~G. Bellemare, Sriram Srinivasan, Georg Ostrovski, Tom Schaul, David
  Saxton, and R{\'{e}}mi Munos.
\newblock Unifying count-based exploration and intrinsic motivation.
\newblock In \emph{Advances in Neural Information Processing Systems 29: Annual
  Conference on Neural Information Processing Systems 2016, December 5-10,
  2016, Barcelona, Spain}, pages 1471--1479, 2016.
\newblock URL
  \url{http://papers.nips.cc/paper/6383-unifying-count-based-exploration-and-intrinsic-motivation}.

\bibitem[Pathak et~al.(2017)Pathak, Agrawal, Efros, and
  Darrell]{pathak2017curiosity}
Deepak Pathak, Pulkit Agrawal, Alexei~A. Efros, and Trevor Darrell.
\newblock Curiosity-driven exploration by self-supervised prediction.
\newblock In Doina Precup and Yee~Whye Teh, editors, \emph{Proceedings of the
  34th International Conference on Machine Learning}, volume~70 of
  \emph{Proceedings of Machine Learning Research}, pages 2778--2787,
  International Convention Centre, Sydney, Australia, 06--11 Aug 2017. PMLR.
\newblock URL \url{http://proceedings.mlr.press/v70/pathak17a.html}.

\bibitem[Ostrovski et~al.(2017)Ostrovski, Bellemare, van~den Oord, and
  Munos]{bellemare_density}
Georg Ostrovski, Marc~G. Bellemare, A{\"{a}}ron van~den Oord, and R{\'{e}}mi
  Munos.
\newblock Count-based exploration with neural density models.
\newblock In \emph{Proceedings of the 34th International Conference on Machine
  Learning, {ICML} 2017, Sydney, NSW, Australia, 6-11 August 2017}, pages
  2721--2730, 2017.
\newblock URL \url{http://proceedings.mlr.press/v70/ostrovski17a.html}.

\bibitem[Machado et~al.(2018)Machado, Bellemare, and Bowling]{machado2018count}
Marlos~C Machado, Marc~G Bellemare, and Michael Bowling.
\newblock Count-based exploration with the successor representation.
\newblock \emph{arXiv preprint arXiv:1807.11622}, 2018.

\bibitem[Bachman et~al.(2018)Bachman, Islam, Sordoni, and
  Ahmed]{bachman2018vfunc}
Philip Bachman, Riashat Islam, Alessandro Sordoni, and Zafarali Ahmed.
\newblock Vfunc: a deep generative model for functions.
\newblock \emph{arXiv preprint arXiv:1807.04106}, 2018.

\bibitem[Goyal et~al.(2019)Goyal, Islam, Strouse, Ahmed, Botvinick, Larochelle,
  Levine, and Bengio]{infobot}
Anirudh Goyal, Riashat Islam, Daniel Strouse, Zafarali Ahmed, Matthew
  Botvinick, Hugo Larochelle, Sergey Levine, and Yoshua Bengio.
\newblock Infobot: Transfer and exploration via the information bottleneck.
\newblock \emph{International Conference on Learning Representations},
  abs/1901.10902, 2019.
\newblock URL \url{http://arxiv.org/abs/1901.10902}.

\bibitem[Eysenbach et~al.(2019)Eysenbach, Gupta, Ibarz, and
  Levine]{eysenbach2019diversity}
Benjamin Eysenbach, Abhishek Gupta, Julian Ibarz, and Sergey Levine.
\newblock Diversity is all you need: Learning skills without a reward function.
\newblock \emph{International Conference on Learning Representations}, 2019.

\bibitem[Pong et~al.(2019)Pong, Dalal, Lin, Nair, Bahl, and
  Levine]{pong2019skew}
Vitchyr~H Pong, Murtaza Dalal, Steven Lin, Ashvin Nair, Shikhar Bahl, and
  Sergey Levine.
\newblock Skew-fit: State-covering self-supervised reinforcement learning.
\newblock \emph{arXiv preprint arXiv:1903.03698}, 2019.

\bibitem[Sutton et~al.(1999)Sutton, McAllester, Singh, and Mansour]{sutton}
Richard~S. Sutton, David~A. McAllester, Satinder~P. Singh, and Yishay Mansour.
\newblock Policy gradient methods for reinforcement learning with function
  approximation.
\newblock In \emph{Advances in Neural Information Processing Systems 12,
  {[NIPS} Conference, Denver, Colorado, USA, November 29 - December 4, 1999]},
  pages 1057--1063, 1999.
\newblock URL
  \url{http://papers.nips.cc/paper/1713-policy-gradient-methods-for-reinforcement-learning-with-function-approximation}.

\bibitem[Schulman et~al.(2017{\natexlab{a}})Schulman, Chen, and
  Abbeel]{schulman2017equivalence}
John Schulman, Xi~Chen, and Pieter Abbeel.
\newblock Equivalence between policy gradients and soft q-learning.
\newblock \emph{arXiv preprint arXiv:1704.06440}, 2017{\natexlab{a}}.

\bibitem[Kingma and Welling(2014)]{vae_kingma}
Diederik~P. Kingma and Max Welling.
\newblock Auto-encoding variational bayes.
\newblock In \emph{2nd International Conference on Learning Representations,
  {ICLR} 2014, Banff, AB, Canada, April 14-16, 2014, Conference Track
  Proceedings}, 2014.
\newblock URL \url{http://arxiv.org/abs/1312.6114}.

\bibitem[Brockman et~al.(2016)Brockman, Cheung, Pettersson, Schneider,
  Schulman, Tang, and Zaremba]{brockman2016openai}
Greg Brockman, Vicki Cheung, Ludwig Pettersson, Jonas Schneider, John Schulman,
  Jie Tang, and Wojciech Zaremba.
\newblock Openai gym.
\newblock \emph{arXiv preprint arXiv:1606.01540}, 2016.

\bibitem[Schaul et~al.(2015)Schaul, Horgan, Gregor, and Silver]{Schaul}
Tom Schaul, Daniel Horgan, Karol Gregor, and David Silver.
\newblock Universal value function approximators.
\newblock In \emph{Proceedings of the 32nd International Conference on Machine
  Learning, {ICML} 2015, Lille, France, 6-11 July 2015}, pages 1312--1320,
  2015.
\newblock URL \url{http://jmlr.org/proceedings/papers/v37/schaul15.html}.

\bibitem[Williams(1992)]{reinforce}
Ronald~J. Williams.
\newblock Simple statistical gradient-following algorithms for connectionist
  reinforcement learning.
\newblock \emph{Machine Learning}, 8:\penalty0 229--256, 1992.
\newblock \doi{10.1007/BF00992696}.
\newblock URL \url{https://doi.org/10.1007/BF00992696}.

\bibitem[Konda and Tsitsiklis(2000)]{konda2000actor}
Vijay~R Konda and John~N Tsitsiklis.
\newblock Actor-critic algorithms.
\newblock In \emph{Advances in neural information processing systems}, pages
  1008--1014, 2000.

\bibitem[Schulman et~al.(2015)Schulman, Moritz, Levine, Jordan, and
  Abbeel]{schulman2015high}
John Schulman, Philipp Moritz, Sergey Levine, Michael Jordan, and Pieter
  Abbeel.
\newblock High-dimensional continuous control using generalized advantage
  estimation.
\newblock \emph{arXiv preprint arXiv:1506.02438}, 2015.

\bibitem[Chevalier-Boisvert et~al.(2018)Chevalier-Boisvert, Willems, and
  Pal]{gym_minigrid}
Maxime Chevalier-Boisvert, Lucas Willems, and Suman Pal.
\newblock Minimalistic gridworld environment for openai gym.
\newblock \url{https://github.com/maximecb/gym-minigrid}, 2018.

\bibitem[Chevalier-Boisvert(2018)]{gym_miniworld}
Maxime Chevalier-Boisvert.
\newblock gym-miniworld environment for openai gym.
\newblock \url{https://github.com/maximecb/gym-miniworld}, 2018.

\bibitem[Schulman et~al.(2017{\natexlab{b}})Schulman, Wolski, Dhariwal,
  Radford, and Klimov]{PPO}
John Schulman, Filip Wolski, Prafulla Dhariwal, Alec Radford, and Oleg Klimov.
\newblock Proximal policy optimization algorithms.
\newblock \emph{CoRR}, abs/1707.06347, 2017{\natexlab{b}}.
\newblock URL \url{http://arxiv.org/abs/1707.06347}.

\bibitem[Todorov et~al.(2012)Todorov, Erez, and Tassa]{conf/iros/TodorovET12}
Emanuel Todorov, Tom Erez, and Yuval Tassa.
\newblock Mujoco: A physics engine for model-based control.
\newblock In \emph{IROS}, pages 5026--5033. IEEE, 2012.
\newblock ISBN 978-1-4673-1737-5.
\newblock URL
  \url{http://dblp.uni-trier.de/db/conf/iros/iros2012.html#TodorovET12}.

\bibitem[Lillicrap et~al.(2016)Lillicrap, Hunt, Pritzel, Heess, Erez, Tassa,
  Silver, and Wierstra]{DDPG}
Timothy~P. Lillicrap, Jonathan~J. Hunt, Alexander Pritzel, Nicolas Heess, Tom
  Erez, Yuval Tassa, David Silver, and Daan Wierstra.
\newblock Continuous control with deep reinforcement learning.
\newblock In \emph{4th International Conference on Learning Representations,
  {ICLR} 2016, San Juan, Puerto Rico, May 2-4, 2016, Conference Track
  Proceedings}, 2016.
\newblock URL \url{http://arxiv.org/abs/1509.02971}.

\bibitem[Henderson et~al.(2018)Henderson, Islam, Bachman, Pineau, Precup, and
  Meger]{henderson}
Peter Henderson, Riashat Islam, Philip Bachman, Joelle Pineau, Doina Precup,
  and David Meger.
\newblock Deep reinforcement learning that matters.
\newblock In \emph{Proceedings of the Thirty-Second {AAAI} Conference on
  Artificial Intelligence, (AAAI-18), the 30th innovative Applications of
  Artificial Intelligence (IAAI-18), and the 8th {AAAI} Symposium on
  Educational Advances in Artificial Intelligence (EAAI-18), New Orleans,
  Louisiana, USA, February 2-7, 2018}, pages 3207--3214, 2018.
\newblock URL
  \url{https://www.aaai.org/ocs/index.php/AAAI/AAAI18/paper/view/16669}.

\bibitem[Ng et~al.(1999)Ng, Harada, and Russell]{NgRewardShaping}
Andrew~Y. Ng, Daishi Harada, and Stuart~J. Russell.
\newblock Policy invariance under reward transformations: Theory and
  application to reward shaping.
\newblock In \emph{Proceedings of the Sixteenth International Conference on
  Machine Learning {(ICML} 1999), Bled, Slovenia, June 27 - 30, 1999}, pages
  278--287, 1999.

\bibitem[Tang et~al.(2017)Tang, Houthooft, Foote, Stooke, Chen, Duan, Schulman,
  DeTurck, and Abbeel]{tang2017exploration}
Haoran Tang, Rein Houthooft, Davis Foote, Adam Stooke, OpenAI~Xi Chen, Yan
  Duan, John Schulman, Filip DeTurck, and Pieter Abbeel.
\newblock \# exploration: A study of count-based exploration for deep
  reinforcement learning.
\newblock In \emph{Advances in neural information processing systems}, pages
  2753--2762, 2017.

\bibitem[Galashov et~al.(2019)Galashov, Jayakumar, Hasenclever, Tirumala,
  Schwarz, Desjardins, Czarnecki, Teh, Pascanu, and
  Heess]{galashov2019information}
Alexandre Galashov, Siddhant~M Jayakumar, Leonard Hasenclever, Dhruva Tirumala,
  Jonathan Schwarz, Guillaume Desjardins, Wojciech~M Czarnecki, Yee~Whye Teh,
  Razvan Pascanu, and Nicolas Heess.
\newblock Information asymmetry in kl-regularized rl.
\newblock \emph{International Conference on Learning Representations}, 2019.
\newblock \doi{arXiv:1905.01240}.
\newblock URL \url{https://arxiv.org/abs/1905.01240}.

\bibitem[Goyal et~al.(2018)Goyal, Brakel, Fedus, Lillicrap, Levine, Larochelle,
  and Bengio]{recall}
Anirudh Goyal, Philemon Brakel, William Fedus, Timothy~P. Lillicrap, Sergey
  Levine, Hugo Larochelle, and Yoshua Bengio.
\newblock Recall traces: Backtracking models for efficient reinforcement
  learning.
\newblock \emph{CoRR}, abs/1804.00379, 2018.
\newblock URL \url{http://arxiv.org/abs/1804.00379}.

\bibitem[Kostrikov(2018)]{pytorchrl}
Ilya Kostrikov.
\newblock Pytorch implementations of reinforcement learning algorithms.
\newblock \url{https://github.com/ikostrikov/pytorch-a2c-ppo-acktr-gail}, 2018.

\bibitem[Fujimoto et~al.(2018)Fujimoto, van Hoof, and Meger]{td3}
Scott Fujimoto, Herke van Hoof, and David Meger.
\newblock Addressing function approximation error in actor-critic methods.
\newblock In \emph{Proceedings of the 35th International Conference on Machine
  Learning, {ICML} 2018, Stockholmsm{\"{a}}ssan, Stockholm, Sweden, July 10-15,
  2018}, pages 1582--1591, 2018.
\newblock URL \url{http://proceedings.mlr.press/v80/fujimoto18a.html}.

\end{thebibliography}
\bibliographystyle{unsrtnat}

\appendix
\renewcommand\thefigure{\thesection\arabic{figure}}    
\setcounter{figure}{0}
\setcounter{section}{18}

\section{Supplementary Material: Marginalized State Distribution Entropy Regularization in Policy Optimization}

\subsection{Additional Experimental Results}

\begin{figure}[h]
\begin{center}
    \begin{subfigure}[b]{0.48\columnwidth}
    \centering
        \includegraphics[width=1.\columnwidth]{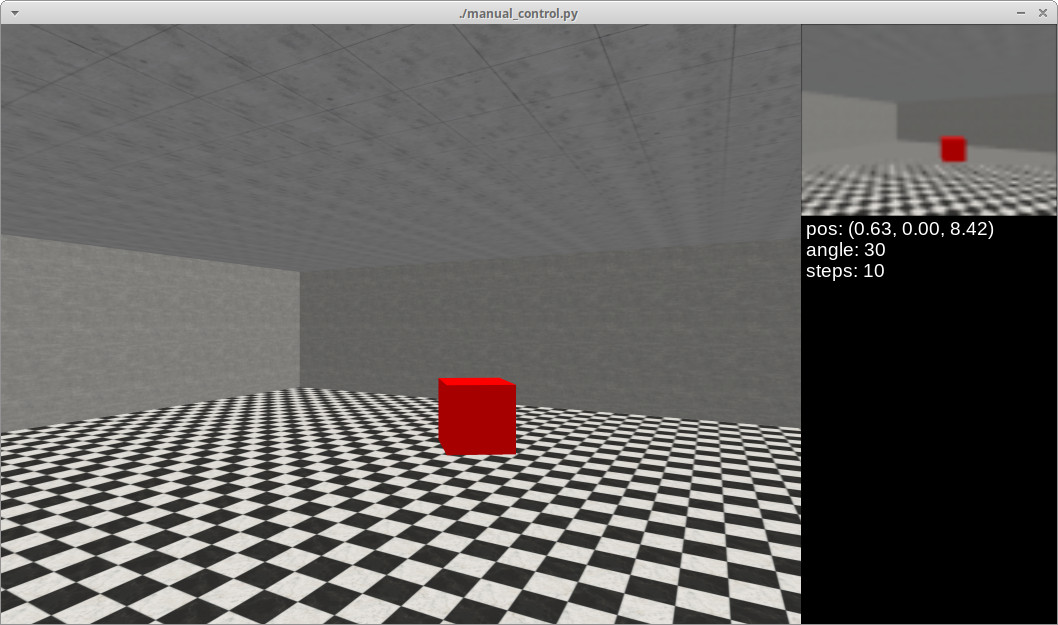}
    \caption{One Room Maze Domain (3D)}
    \end{subfigure}
    \begin{subfigure}[b]{0.48\columnwidth}
    \centering
        \includegraphics[width=1.\columnwidth]{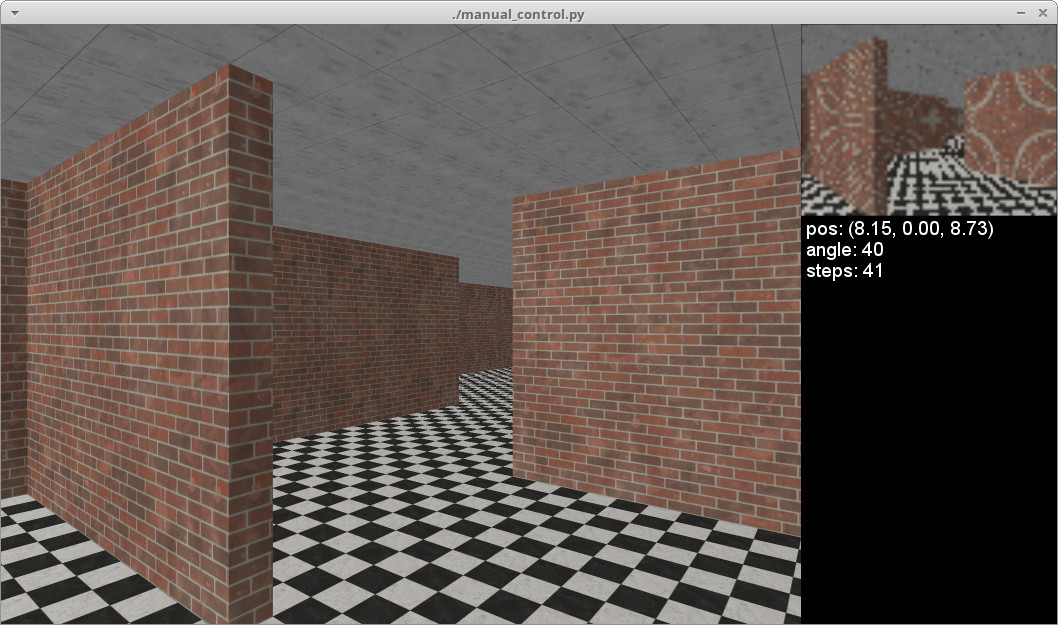}
    \caption{Maze Domain (3D)}
    \end{subfigure}
    \caption{Examples of MiniWorld environments. \label{fig:miniworld_env_plots}}
\end{center}    

\end{figure}

\begin{figure}[t]
\begin{center}
    \begin{subfigure}[b]{0.45\columnwidth}
    \centering
        \includegraphics[width=1.\columnwidth]{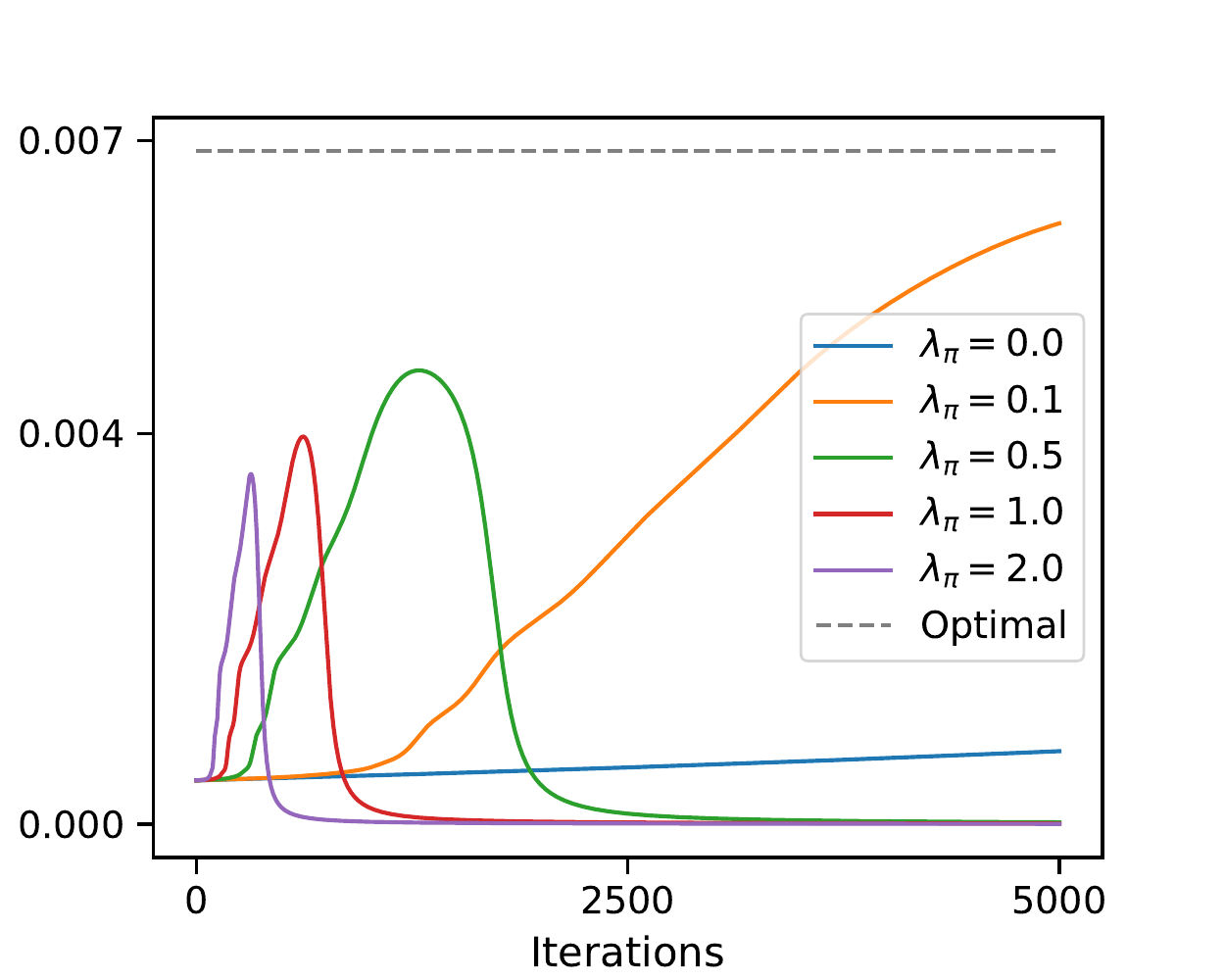}
        \caption{$\lambda_s=0$\label{fig:frozen_lake_policy_ent}}
    \end{subfigure}
    \begin{subfigure}[b]{0.45\columnwidth}
        \centering
        \includegraphics[width=1.\columnwidth]{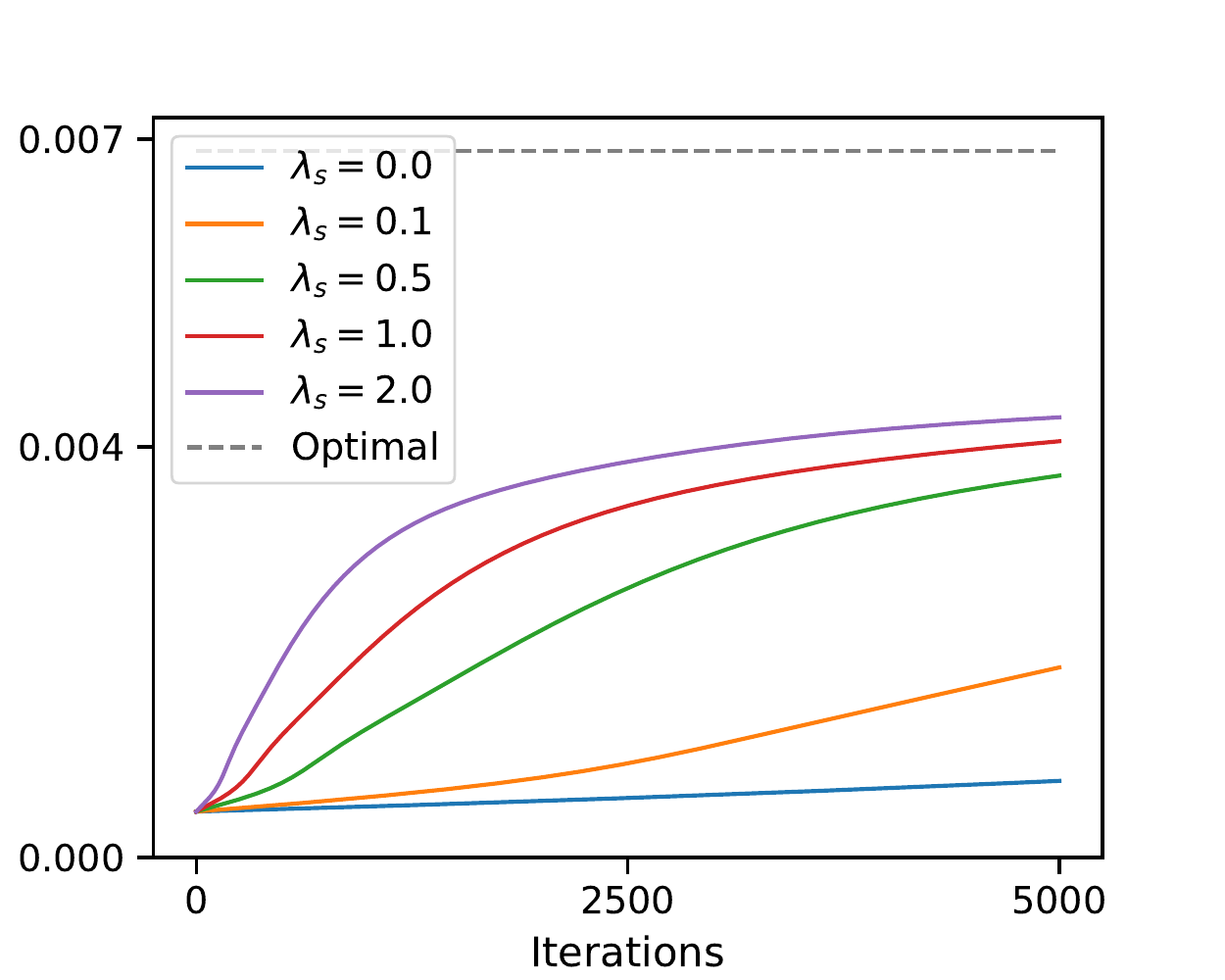}
        \caption{$\lambda_\pi=0$\label{fig:frozen_lake_state_ent}}
    \end{subfigure}
    \begin{subfigure}[b]{0.45\columnwidth}
        \centering
        \includegraphics[width=1.\columnwidth]{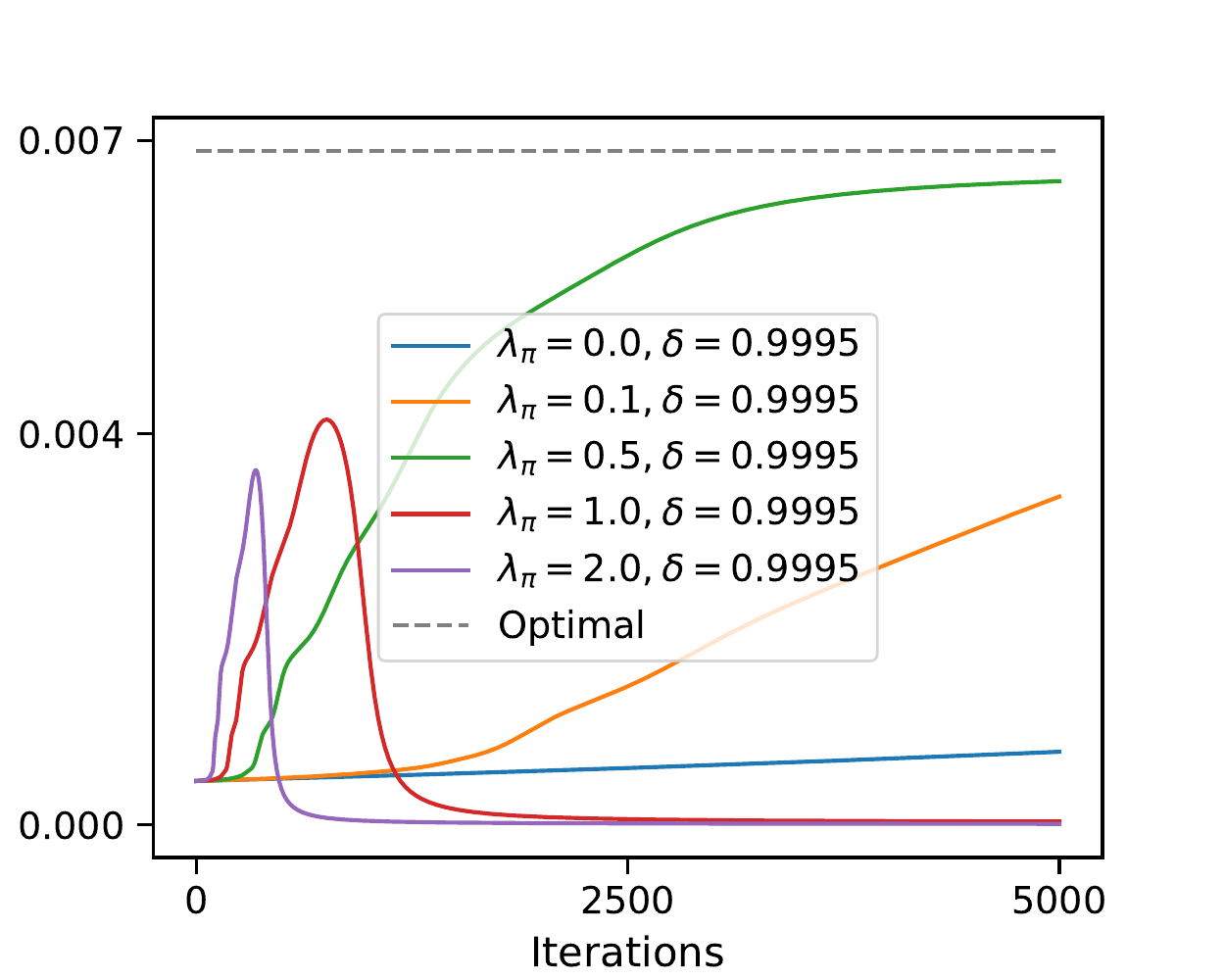}
        \caption{$\lambda_s=0$ with decay on $\lambda_\pi$\label{fig:frozen_lake_pol_ent_w_decay}}
    \end{subfigure}
    \begin{subfigure}[b]{0.45\columnwidth}
        \centering
        \includegraphics[width=1.\columnwidth]{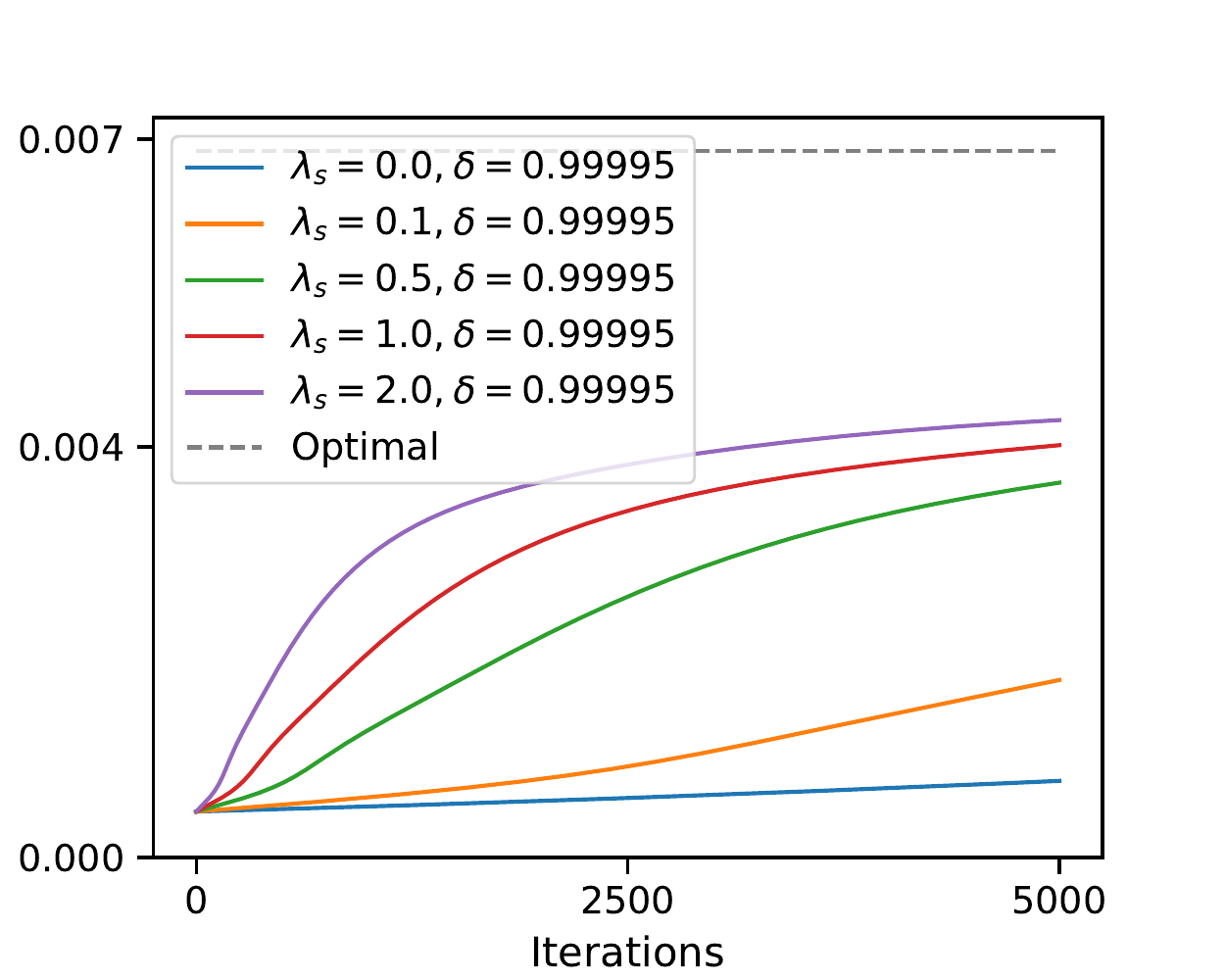}
        \caption{$\lambda_\pi=0$ with decay on $\lambda_s$\label{fig:frozen_lake_state_ent_w_decay}}
    \end{subfigure}
    
    \caption{\small (a)-(d) The full range of experiments on FrozenLake that systematically study the impact of the two kinds of entropy with and without a decay on their regularization coefficients.
    \label{fig:frozen_lake_experiments_full}
    }
\end{center}    
\end{figure}

\begin{figure}[t]
\begin{center}
    \begin{subfigure}[b]{0.32\columnwidth}
    \centering
        \includegraphics[width=1.\columnwidth]{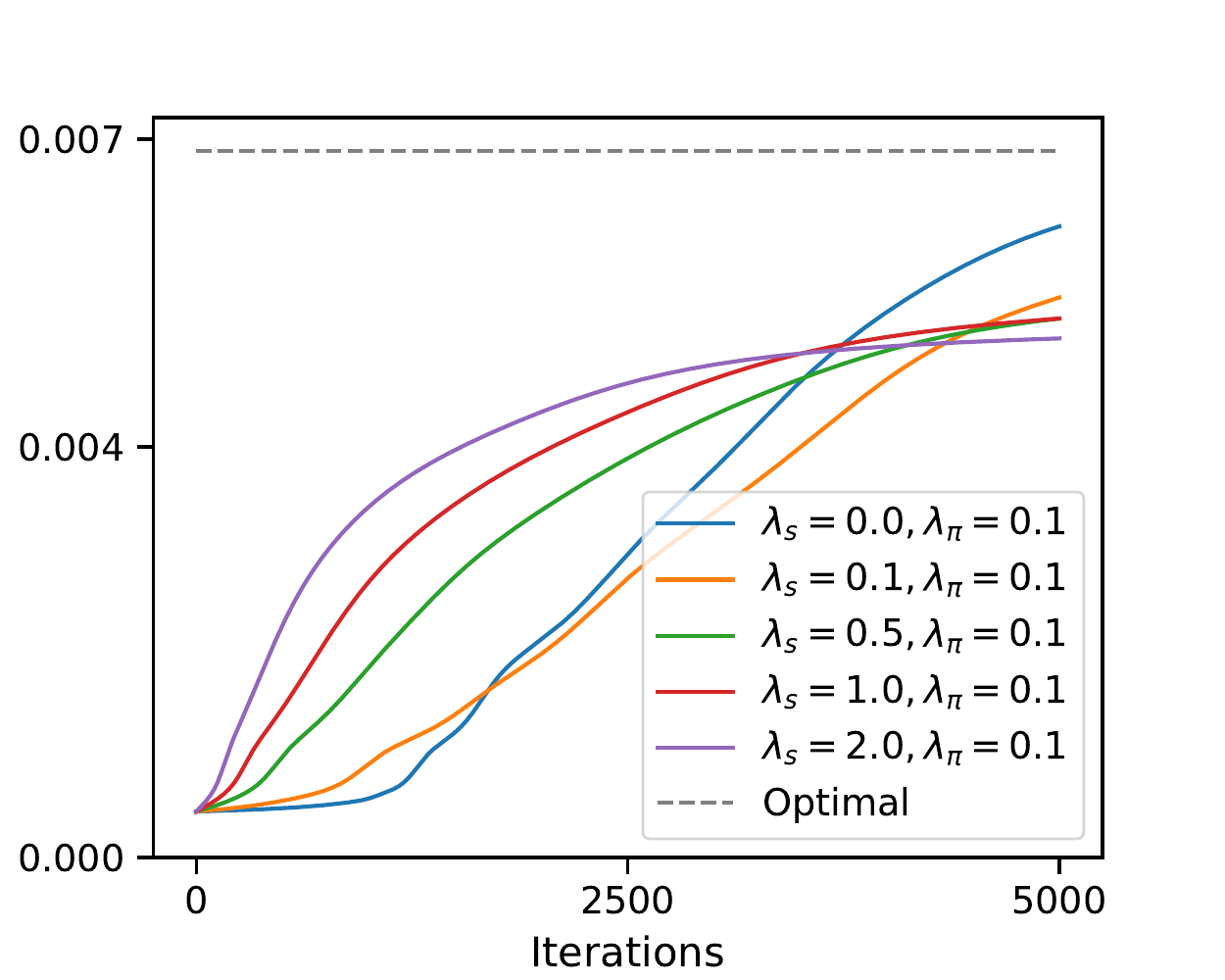}
        \caption{$\lambda_\pi=0.1$\label{fig:frozen_lake_interact_pol01}}
    \end{subfigure}
    \begin{subfigure}[b]{0.32\columnwidth}
        \centering
        \includegraphics[width=1.\columnwidth]{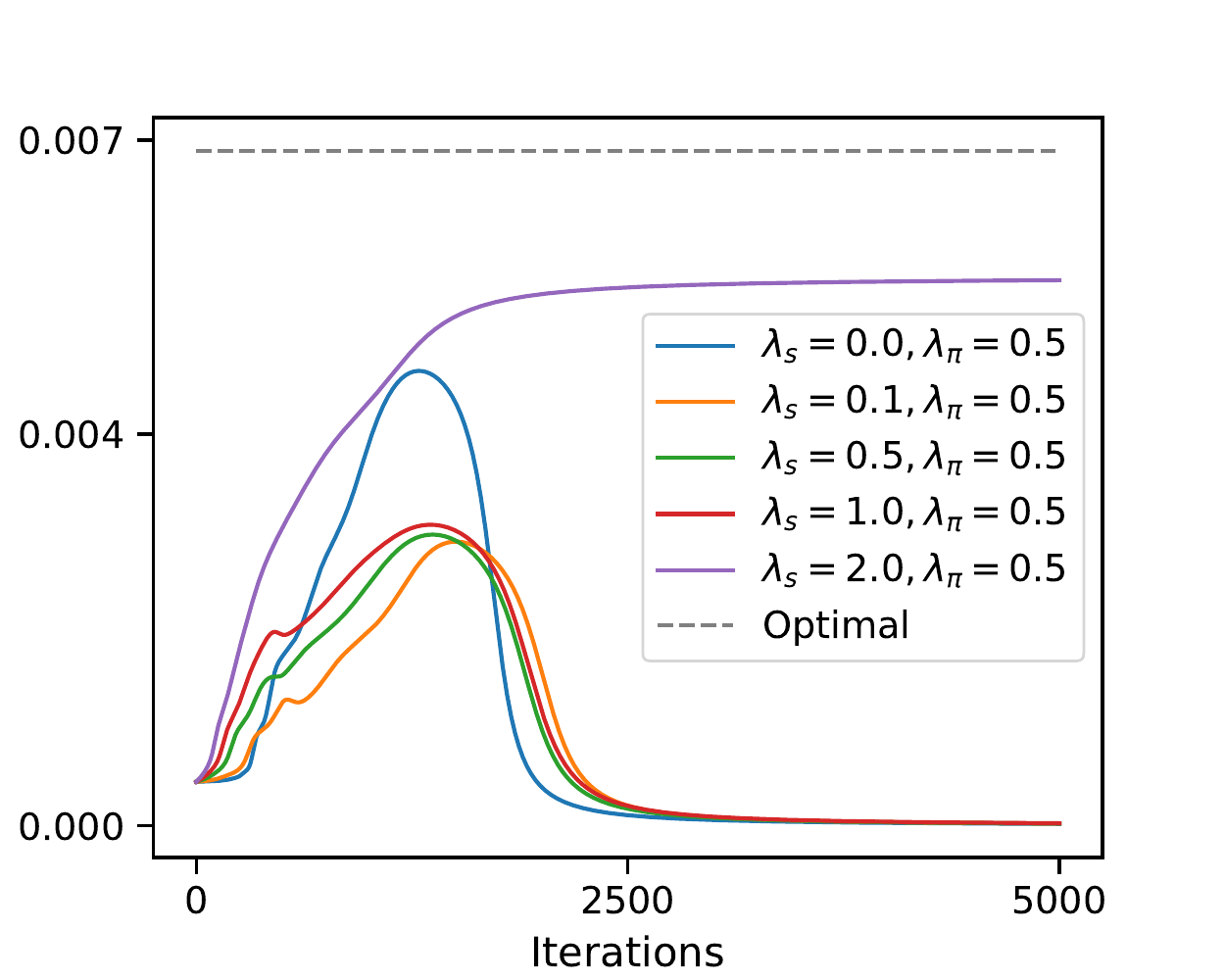}
        \caption{$\lambda_s=0.5$\label{fig:frozen_lake_interact_pol05}}
    \end{subfigure}
    \begin{subfigure}[b]{0.32\columnwidth}
        \centering
        \includegraphics[width=1.\columnwidth]{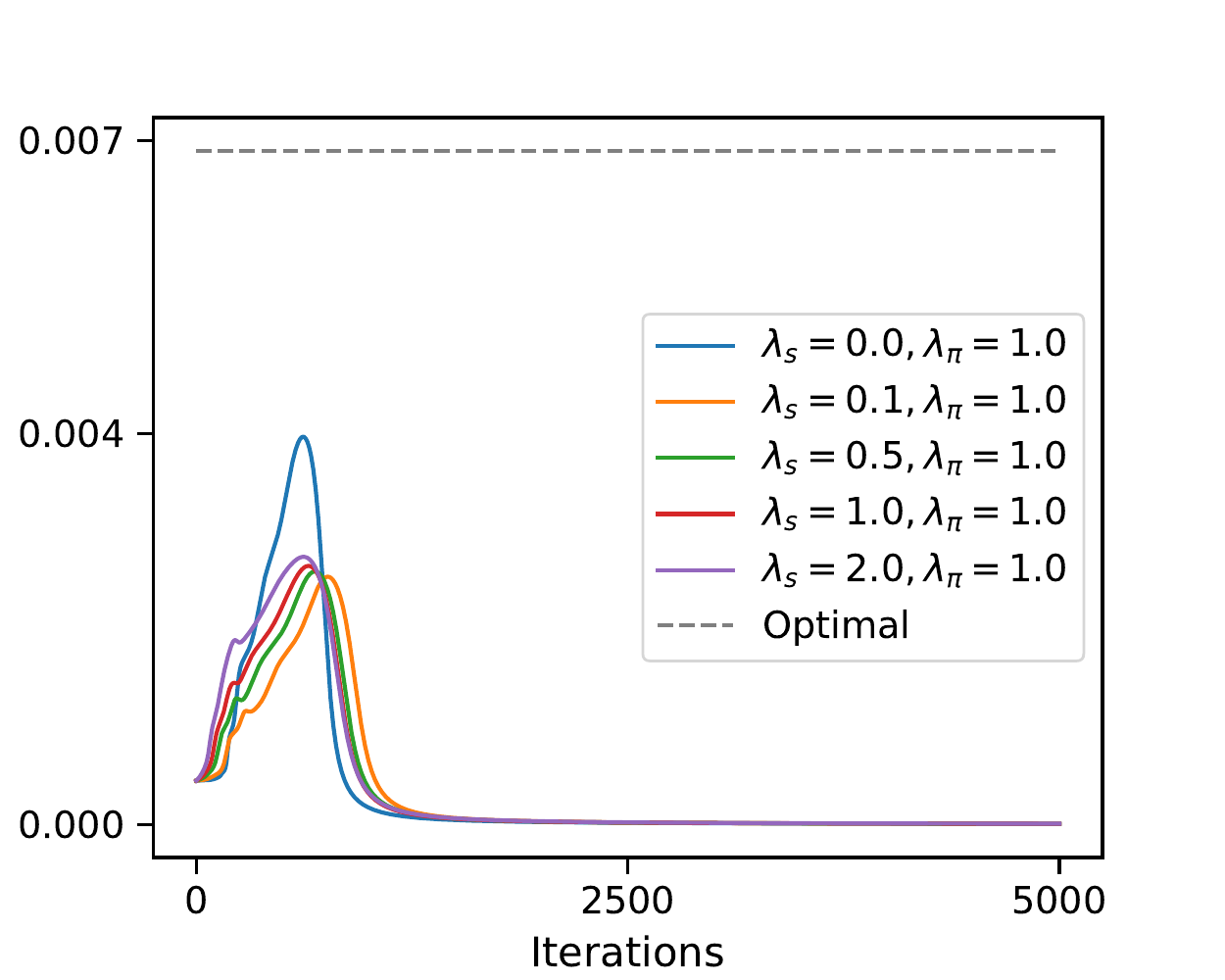}
        \caption{$\lambda_s=1.0$\label{fig:frozen_lake_interact_pol10}}
    \end{subfigure}
    \caption{\small (a)-(c) The full range of experiments on FrozenLake that systematically study the impact of the interaction between two kinds of entropy.
    \label{fig:frozen_lake_experiments_full_interaction}
    }
\end{center}    
\end{figure}

\begin{figure}[ht]
\begin{center}
    \begin{subfigure}[b]{0.45\columnwidth}
    \centering
        \includegraphics[width=1.\columnwidth]{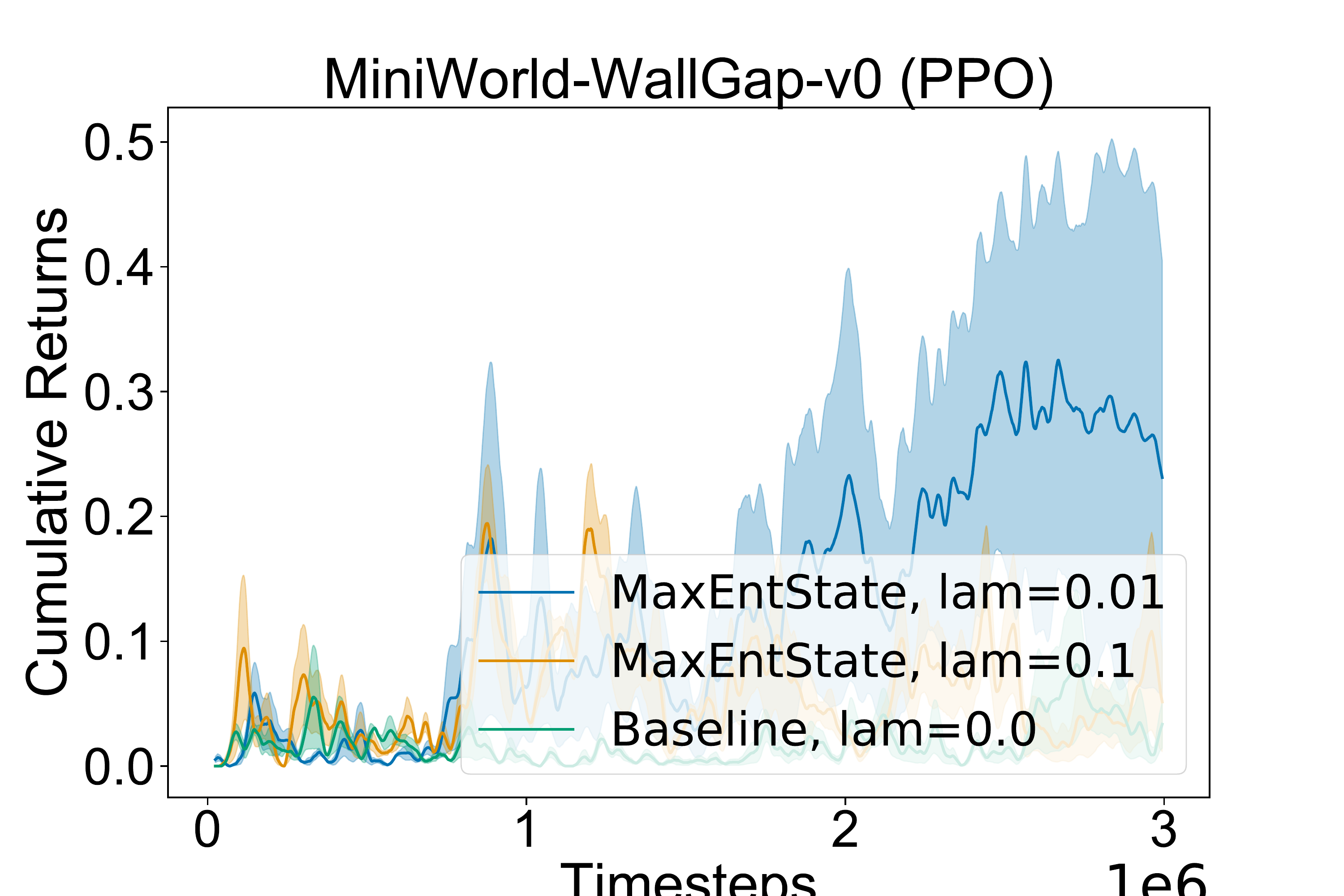}
        \caption{\label{miniworld_wallgap_ppo}}
    \end{subfigure}
    \begin{subfigure}[b]{0.45\columnwidth}
    \centering
        \includegraphics[width=1.\columnwidth]{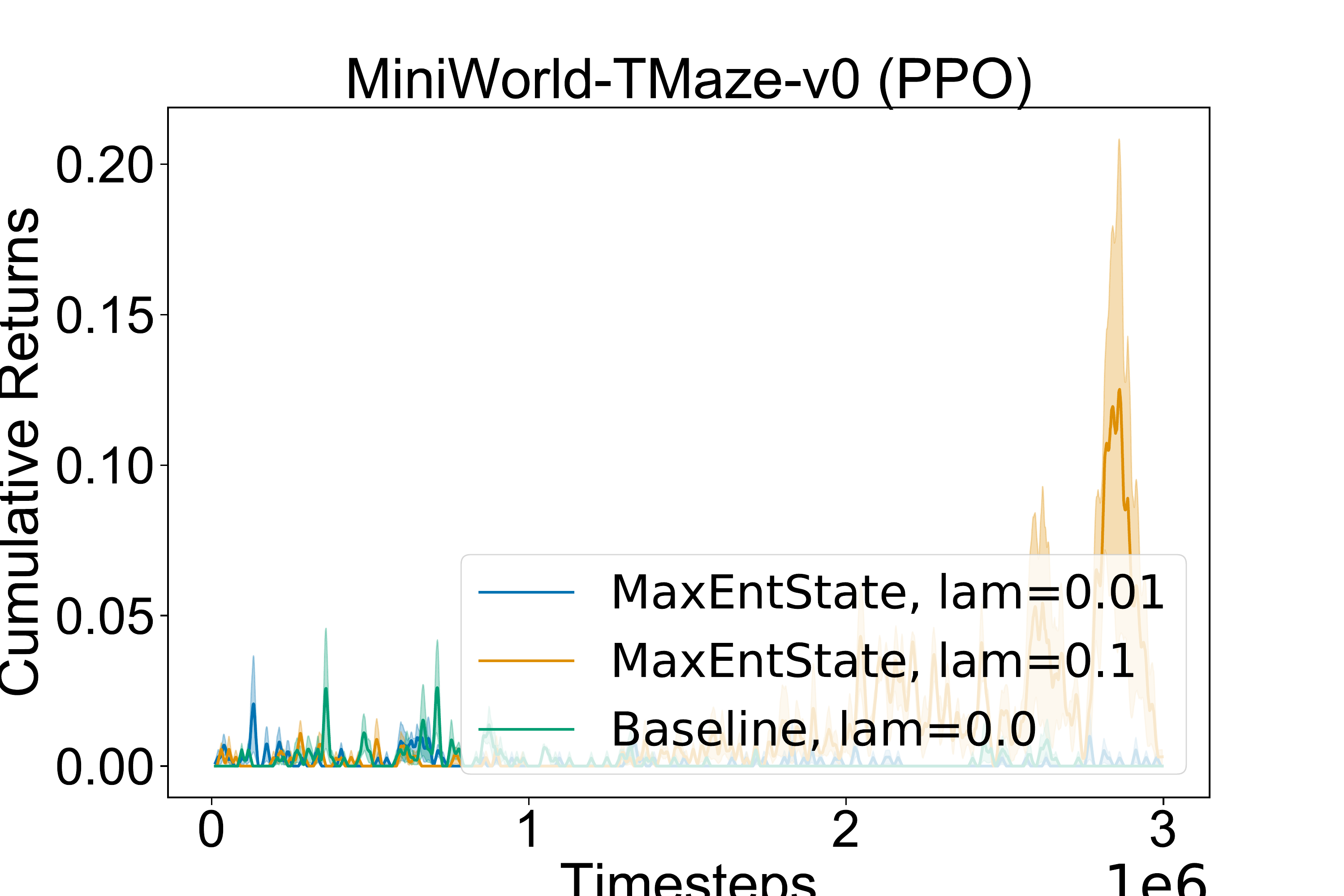}
        \caption{\label{miniworld_tmaze_ppo}}
    \end{subfigure}
    \caption{MiniWorld Envoronments with PPO, with different $\lambda$ weightings for the marginal state entropy regularization. Comparison with standard PPO baseline with policy entropy regularization. We find that with MaxEntState regularization, PPO can solve these hard exploration 3D maze navigation tasks much faster than baseline with MaxEntPolicy.}
    \label{fig:miniworld_ppo}
\end{center}    
\end{figure}

\begin{figure}
\begin{center}
    \begin{subfigure}[b]{0.32\columnwidth}
    \centering
        \includegraphics[width=1.\columnwidth]{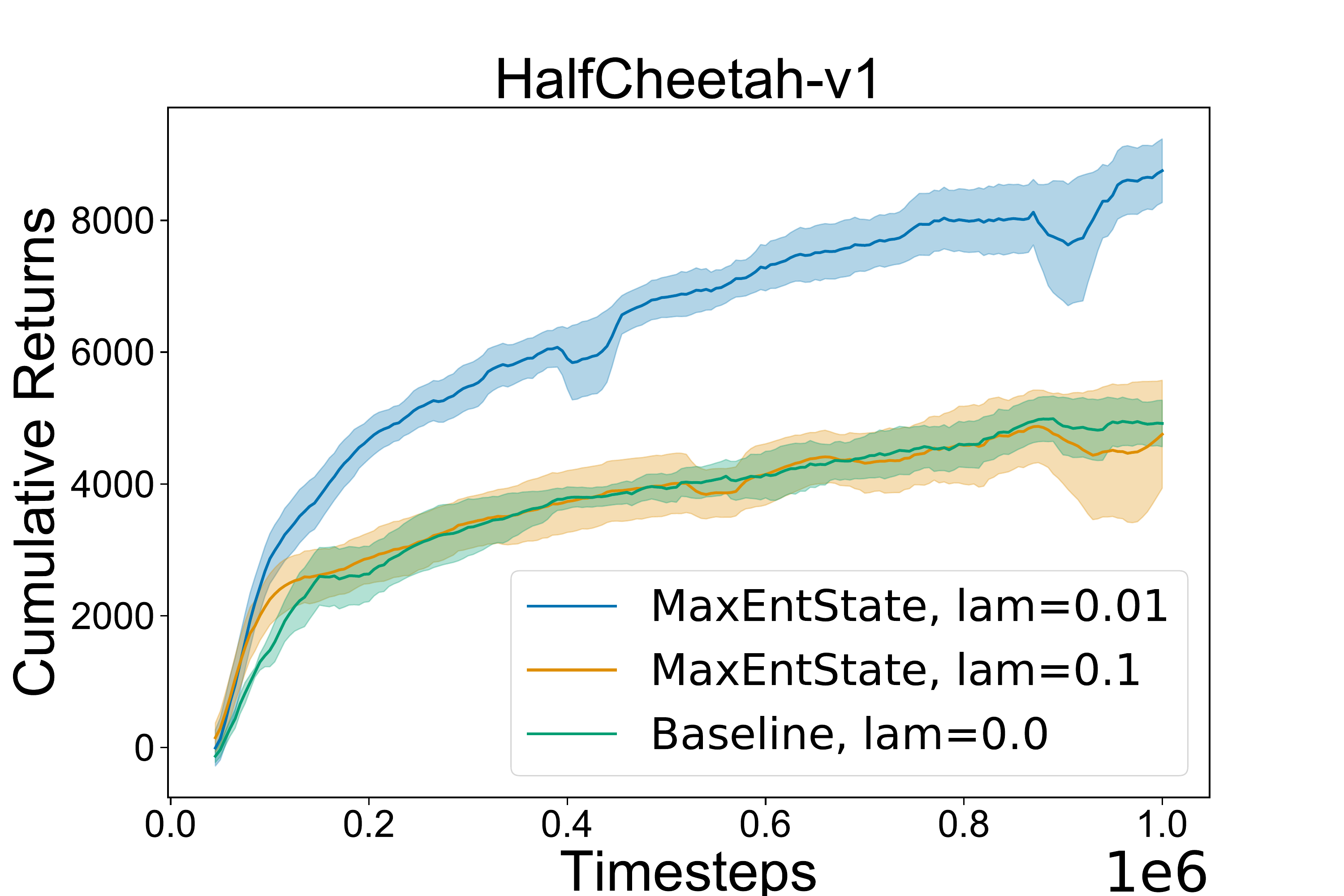}
        \caption{\label{ddpg_halfcheetah}}
    \end{subfigure}
    \begin{subfigure}[b]{0.32\columnwidth}
    \centering
        \includegraphics[width=1.\columnwidth]{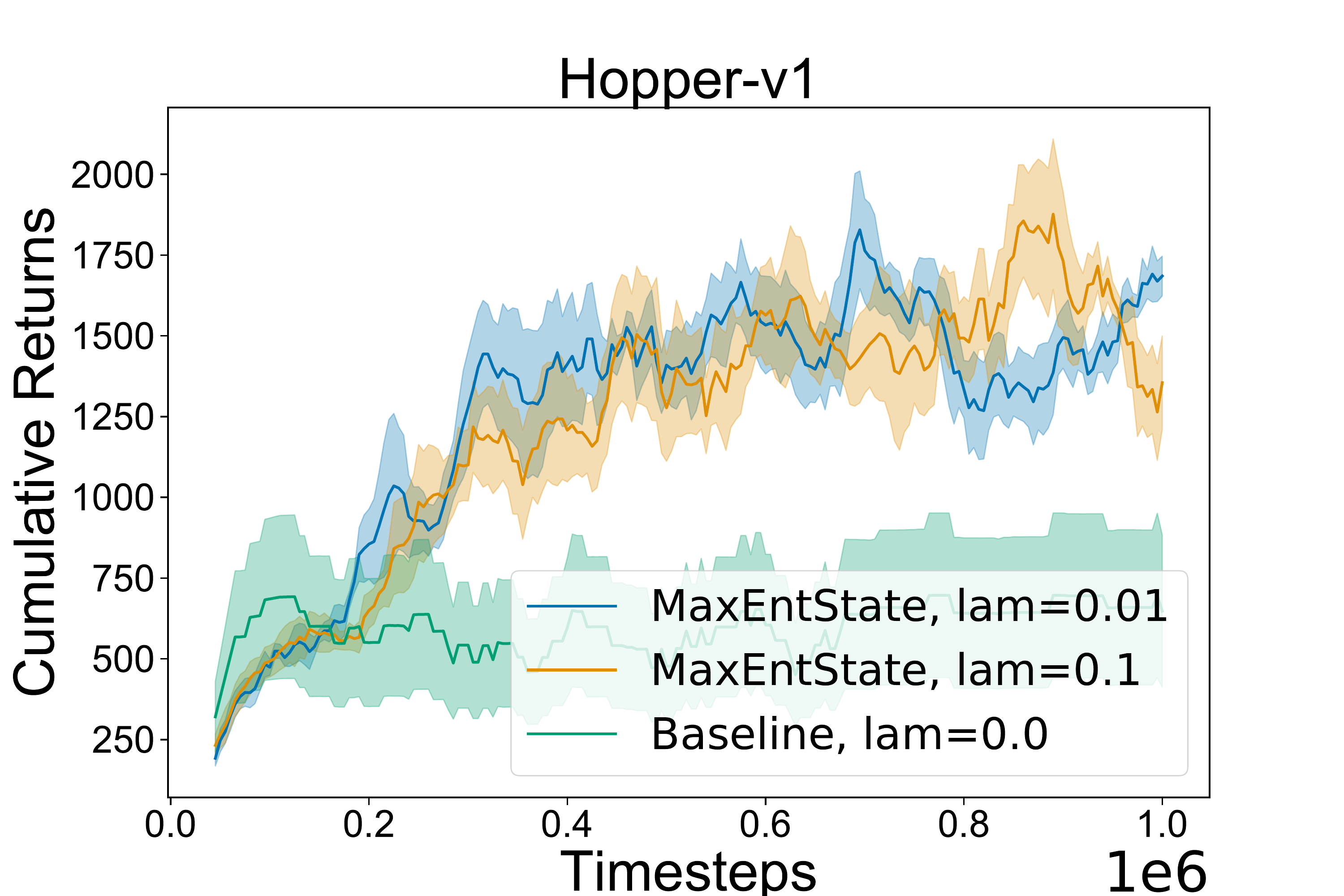}
        \caption{\label{ddpg_hopper}}
    \end{subfigure}
    \quad
    \begin{subfigure}[b]{0.32\columnwidth}
    \centering
        \includegraphics[width=1.\columnwidth]{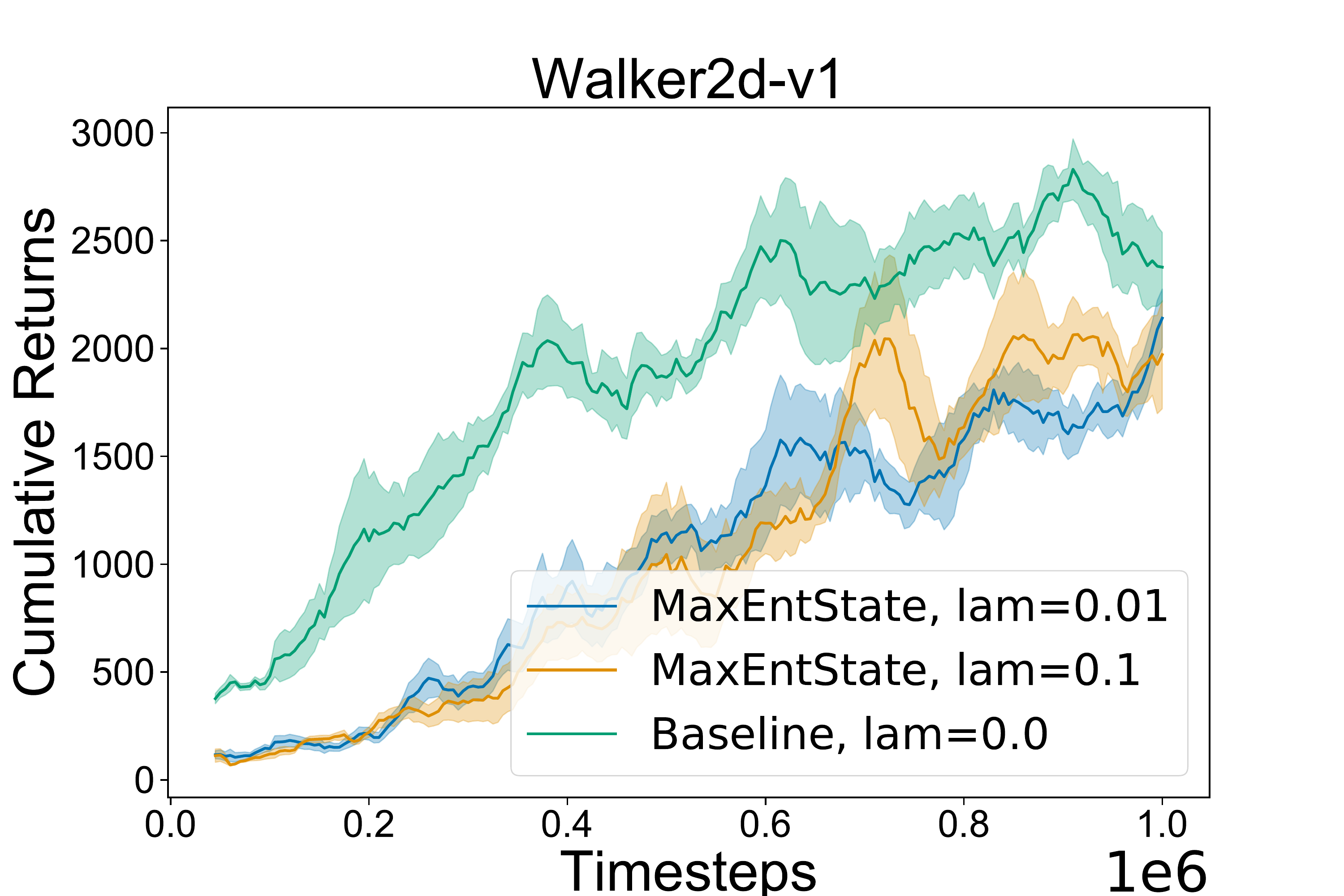}
        \caption{\label{ddpg_walker}}
    \end{subfigure}
    \caption{Continuous Control Benchmarks on DDPG. Comparison with DDPG with and without state entropy regularization. We find that with MaxEntState, we can significantly improve the performance of DDPG, since this algorithm cannot maximize entropy with deterministic policies. Maximizing entropy with the marginal state distribution therefore plays a key role with exploration, improving the sample efficiency of this off-policy algorithm.}
    \label{fig:DDPG}
\end{center}    
\end{figure}

\begin{figure}[h]
\begin{center}
    \begin{subfigure}[b]{0.2\columnwidth}
    \centering
        \includegraphics[width=1.\columnwidth]{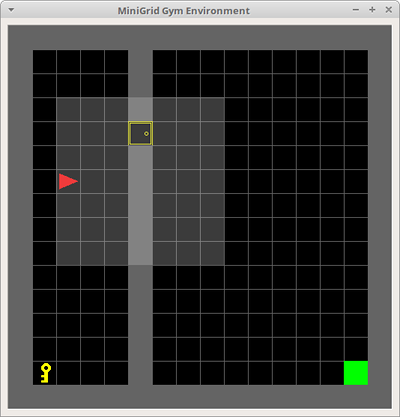}
    \caption{Doorkey Env}
    \end{subfigure}
    \begin{subfigure}[b]{0.2\columnwidth}
    \centering
        \includegraphics[width=1.\columnwidth]{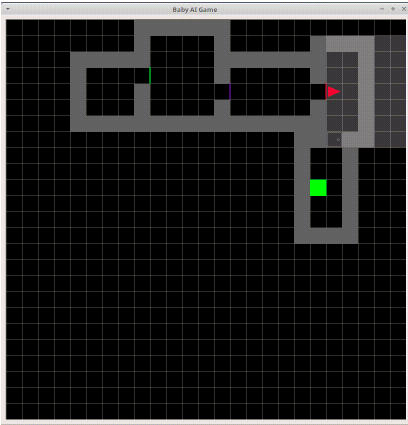}
    \caption{Multi-Room Env}
    \end{subfigure}
    \begin{subfigure}[b]{0.2\columnwidth}
    \centering
        \includegraphics[width=1.\columnwidth]{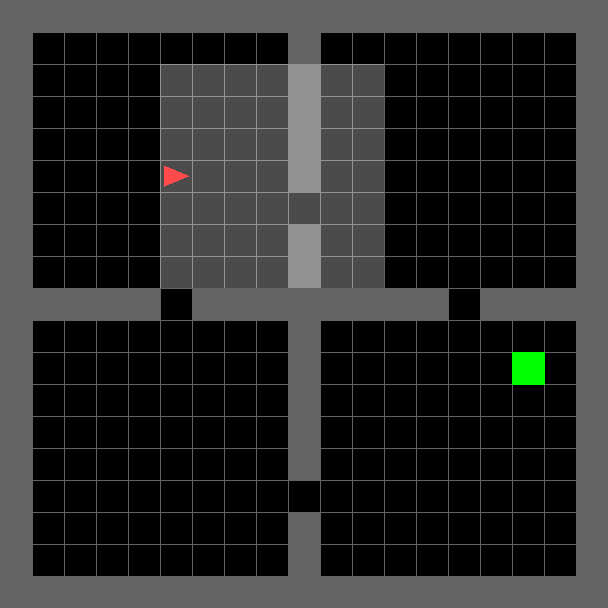}
    \caption{Four Rooms Env}
    \end{subfigure}
    \caption{Examples of MiniGird environments.\label{fig:minigrid_env_plots}} 
\end{center}    
\end{figure}

\subsection{Derivation of State Entropy Regularized Policy Gradient} 
\label{sec:appendix:pg_derivation}
In this section, we derive the policy gradient theorem with the state entropy regularized objective. Since the objective function is given by $V_{\pi}(s_0) = \sum_{a}\pi(a \mid s=s_0) Q_{\pi}(s,a)$, with state entropy regularization, we can write $Q_{\pi}(s,a)$ as follows

\begin{align}
      Q_{\pi}(s,a) = [r(s,a) + \lambda \ent (p_{\pi}(s)) + \gamma \sum_{s'} P(s' \mid s=s_0, a) V_{\pi}(s')]  
\end{align}

We can now derive the policy gradient theorem, with the explicit state entropy regularized objective given by 

\begin{align}
\begin{split}
    \nabla_{\theta} V_{\pi}(s_0) &= \nabla_{\theta} \sum_{a} \pi(a|s) Q_{\pi}(s,a)\\
    &= \sum_{a} Q_{\pi}(s,a) \nabla_{\theta} \pi(a|s) + \sum_{a} \pi(a|s)\nabla_{\theta} Q_{\pi}(s,a)
\end{split}
\end{align}

The gradient of the term $\nabla_{\theta} Q_{\pi}(s,a)$ is : 

\begin{equation}
\begin{split}
    \nabla_{\theta} Q_{\pi}(s,a) &= \nabla_{\theta} \Bigl[  r(s,a) + \lambda \ent (p_{\pi}(s)) + \gamma \sum_{s'} P(s' \mid s, a) V_{\pi}(s')\bigr]\\
    &= \nabla_{\theta} \lambda \ent (p_{\pi}(s)) + \gamma \sum_{s'} P(s' \mid s, a) \nabla_{\theta} V_{\pi}(s')
\end{split}
\end{equation}

Following the policy gradient theorem \citep{sutton}, and using the log-derivative trick, we can therefore write

\begin{equation}
    \nabla_{\theta} V_{\pi}(s) = \sum_{a} \pi(a|s) \Bigl[ Q_{\pi}(s,a) \nabla_{\theta} \log \pi(a|s) + \nabla_{\theta} \lambda \ent (p_{\pi}(s)) \Bigr] + \sum_{a} \pi(a|s) \gamma \sum_{s'} P(s' \mid s, a) \nabla_{\theta} V_{\pi}(s')    
\end{equation}

from which we can therefore get the following policy gradient theorem. The key difference here is that, the derived for the entropy of the marginal state distribution gives a regularized policy gradient update, where the original objective can be retrieved with $\lambda=0$.

\begin{equation}
    \nabla_{\theta} J(\theta) = \EXP_{\pi_{\theta}}  \Bigl[ \nabla_{\theta} \log \pi_{\theta}(a | s) Q^{\pi_{\theta},\lambda}(a, s) + \lambda \nabla_{\theta} \ent ( p_{\pi}(s) )     \Bigr]
\end{equation}

In practice, however, computing $p_{\pi}(s)$ exactly is difficult. This is because this term can be interpreted as using a probability density estimate of each state $p(s)$ which is explicitly dependent on the policy $\pi_{\theta}$ which is parameterized.

\subsection{Reproducibility Checklist}

We follow the reproducibility checklist from \href{https://www.cs.mcgill.ca/~jpineau/ReproducibilityChecklist.pdf}{Pineau, 2018} and include further details here. For all the models and algorithms, we have included details that we think would be useful for reproducing the results of this work.

\begin{itemize}

\item For all \textbf{models} and \textbf{algorithms} presented, check if you include:

\begin{enumerate}
    
    \item \textit{Description of Algorithm and Model} : We have included an algorithm box that clearly outlines the proposed approach. Our method can be used on top of any existing policy-gradient based RL algorithm, and in our experiments, we mostly used Reinforce, PPO and A2C algorithms for demonstrations. To implement our approach, we simply need to implement an encoder to the policy network, that maps states to a fixed dimensional latent variable (in most experiments, we use the size of the latent space $Z=64$). We assume a Gaussian distribution over the latent representation, and compute the variational entropy $\ent (q_{\pi}(z|s)))$ and the KL divergence $KL(q_{\phi}(z|s)||p(z))$. We use this as a regularization in the policy gradient update. We further use the entropy $\ent (q_{\pi}(z|s)))$ to provide an exploration bonus in hard exploration tasks. This regularization is added on top of existing max-entropy policy regularization (which is also added as an exploration bonus). This term is added with a $\lambda$ weighting term.

    \item \textit{Analysis of Complexity} : We do not include any separate analysis of the complexity of our algorithm. Computation-wise, our approach requires the extra computation of the KL-term for the regularizer (which is similar to a lot of existing related works).
    
    \item \textit{Link to downloadable source code} : See the experimental details section below, where we include further details for each of our experimental setup. All our implementations use existing open-sourced RL implementations (details of which we list below). We provide the code used in our experiments in a separate zip file, and agree that we will open source our implementations, for any result figures used in this paper, as well as scripts used for launching and plotting the experimental results for absolute clarity.
    
\end{enumerate}

\item For any \textbf{theoretical claim}, check if you include:

\begin{enumerate}
    \item We include a statement of our theoretical result in the main paper. We clearly describe the theoretical steps required to derive our modified objective.
    
    \item \textit{Complete Proof of Claim :} In appendix, we have also included a clear derivation of our proposed approach, using existing theorem used in the literature, to clarify how exactly our proposed approach differs.
    
    \item \textit{A clear explanation of any assumptions} : We clearly describe the assumptions made to make our model work in practice. Since we introduce an encoder in our policy network for practical realization of the algorithm, we clearly mention that the only assumption being made is assuming a standard Gaussian distribution as output of the encoder, and a unit Gaussian prior for computing the KL divergence term.
    
\end{enumerate}

\item For all figures and tables that present \textbf{empirical results}, check if you include:

\begin{enumerate}
    \item \textit{Data collection process} : We did not need to include a complete description of the data collection process. This is because we use any standard RL algorithm, and use the same number of timesteps or episodes typically used for practical implementations. Both our proposed approach and the baseline are trained with the same number of samples.
    
    \item \textit{Downloadable version of environment} : We use open-sourced OpenAI gym environments for most of our experiments, including the Mujoco simulator. For experiments, where we used other environments that are typically not used, we either include a link to the environment code repo that we used, or provide the actual code of the environment in the accompanying codebase of this paper. 
    
    \item \textit{Description of any pre-processing step} : We do not require any data pre-processing step for our experiments. 
    
    \item \textit{Sample allocation for training and evaluation} : We use standard RL evaluation framework for our experimental results. In our experiments, as done in any RL algorithm, the trained policy is evaluated at fixed intervals, and the performance is measured by plotting the cumulative returns. In most of our presented experimental results, we plot the cumulative return performance measure. In 2 of our results, we plot the state visitation heatmap for a qualitative analysis of our method and to present the intuition. In the accompanying code, we also include details of how to generate these heatmaps to reproduce the results.
    
    \item \textit{Range of hyper-parameters considered : }  For our experiments, we did not do any extensive hyperparameter tuning. We took existing implementations of RL algorithms (details of which are given in the Appendix experimental details section below), which generally contain tuned implementations. For our proposed method, we only introduced the extra hyperparameter $\lambda$ for the state entropy weighting. We tried our experiments with only 3 different lambda values ($\lambda = 0.001, 0.01 and 0.1$) and compared to the baseline with $\lambda=0.0$ for a fair comparison. Both our proposed method and the baseline contains the same network architectures, and other hyperparameters, that are used in existing open-sourced RL algorithms. We include more details of our experiment setups in the next section in Appendix.
    
    \item \textit{Number of Experiment Runs} : For all our experimental results, we plot results over $5$ random seeds. Each of our hyper-parameter tuning is also done with $5$ experiment runs with each hyperparameter. These random seeds are sampled at the start of any experiment, and plots are shown averaged over 5 runs. We note that since a lot of DeepRL algorithms suffer from high variance, we therefore have the high variance region in some of our experiment results.
    
    \item \textit{Statistics used to report results} : In the resulting figures, we plot the mean, $\mu$, and standard error $\frac{\sigma}{\sqrt(N)})$ for the shaded region, to demonstrate the variance across runs and around the mean. We note that some of the environments we used in our experiments, are very challenging to solve (e.g 3D maze navigation domains), resulting in the high variance (shaded region) around the plots. The Mujoco control experiments done in this work have the standard shaded region as expected in the performance in the baseline algorithms we have used (DDPG and SAC).
    
    \item \textit{Error bars} : The error bars or shaded region are due to $std / sqrt(N)$ where $N=5$ for the number of experiment runs.
    
    \item \textit{Computing Infrastrucutre} : We used both CPUs and GPUs in all of our experiments, depending on the complexity of the tasks. For some of our experiments, we could have run for more than $5$ random seeds, for each hyperparameter tuning, but it becomes computationally challenging and a waste of resources, for which we limit the number of experiment runs, with both CPU and GPU to be a standrd of $5$ across all setups. 
\end{enumerate}

\end{itemize}

\subsection{Additional Experimental Details}
\label{sec:ref:detailed_experimental_setup}
In this section, we include further experimental details and setup for the results presented in the paper

\textbf{Experiment setup in Demonstrating Hypothesis in Section \ref{sec:res:hazan}}: In section \ref{sec:res:hazan} we demonstrate the effect of maximizing $\ent (d_{\pi})$ on a simple FrozenLake task. we use the standard FrozenLake gym environment \citep{brockman2016openai}, and used a simple actor-critic implementation with one-hot state encoding of states as features, and a one-layer neural network approximator.  
    
\textbf{Experiment setup for State Space Coverage in Section  \ref{sec:res:state_space_coverage}}:
Pachinko world and the double-slit experiment are implemented in the open-source package, EasyMDP\footnote{\href{https://github.com/zafarali/emdp}{https://github.com/zafarali/emdp}
}. For this task, we use a parallel threaded \textsc{Reinforce} implementation, and only compare the performance of our proposed approach qualitatively by plotting the state visitation heatmaps.

For the four rooms domain, we used the open-source four rooms code available from \citep{recall}. We use an actor-critic with a GAE implementation for this task, and compare with and without our proposed state entropy regularization. We provide the code for this task in the accompanying code for the paper.

\textbf{Experiment setup in Maze Navigation Tasks in Section \ref{sec:res:deep_RL}} 
    
For the sparse reward POMDP gridworld tasks, we use the open-sourced Gym-Minigrid environments available in \citep{gym_minigrid}. This implementation uses a standard A2C implementation from \citep{pytorchrl}. We used the same network architectures, learning rates and optimizers, as used in the open-source implementation, with no further hyper-parameter tuning. Figure \ref{fig:minigrid_env_plots} shows the POMDP environments from the Minigrid \citep{gym_minigrid}. For this task, we provide code for our implementation built on top of existing A2C code.

In the 3D maze navigation tasks, we used the 
open-source Miniworld environments from \citep{gym_miniworld} (Figure~\ref{fig:miniworld_env_plots}). For our experiments, we used both A2C and PPO with the open-sourced implementations from \citep{pytorchrl}. We used the same network architecture, optimizers and learning rates for our implementation, as used in the baseline code of \citep{pytorchrl}. Figure below further shows some of the environments used in this work

\textbf{Experiment setup in Continuous Control Tasks in Section \ref{sec:res:continuous_control}}

For the continuous control experiments, we used the open-source implementation of DDPG available from the accompanying paper \citep{td3}. We further use a SAC implementation, from a modified implementation of DDPG. Both the implementations of DDPG and SAC are provided with the accompanying codebase. We used the same architectures and hyperparameters for DDPG and SAC as reported in \citep{td3}.

\end{document}